\documentclass[runningheads]{llncs}

\usepackage{graphicx}
\usepackage{amsmath, amssymb}
\usepackage{breqn}
\usepackage{tabularx}
\usepackage{multirow}
\usepackage{grffile}
\usepackage{color}
\usepackage{float}
\usepackage{url}
\usepackage{lineno}
\usepackage[abs]{overpic}
\usepackage{transparent}
\usepackage{cite}
\usepackage[normalem]{ulem}
\usepackage[colorlinks,linkcolor=red,anchorcolor=blue,citecolor=green]{hyperref}

\usepackage[width=122mm,left=12mm,paperwidth=146mm,height=193mm,top=12mm,paperheight=217mm]{geometry}

\newcommand{\etal}{\textit{et al}}
\begin{document}
%\renewcommand\thelinenumber{\color[rgb]{0.2,0.5,0.8}\normalfont\sffamily\scriptsize\arabic{linenumber}\color[rgb]{0,0,0}}
% \linenumbers
\pagestyle{headings}
\mainmatter
\def\ECCV18SubNumber{1168}  % Insert your submission number here

\title{Shift-Net: Image Inpainting via Deep Feature Rearrangement} % Replace with your title

\titlerunning{Shift-Net: Image Inpainting via Deep Feature Rearrangement}

\authorrunning{Zhaoyi Yan \etal}

\author{Zhaoyi Yan$^{1}$, Xiaoming Li$^{1}$, Mu Li$^{2}$, Wangmeng Zuo$^{1}$\thanks{Corresponding author.}, Shiguang Shan$^{3,4}$\\
{\tt\small yanzhaoyi@outlook.com, csxmli@hit.edu.cn, csmuli@comp.polyu.edu.hk,}\\
{\tt\small wmzuo@hit.edu.cn, sgshan@ict.ac.cn}
\small\institute{$^1$School of Computer Science and Technology, Harbin Institute of Technology, Harbin, China\\
$^2$Department of Computing, The Hong Kong Polytechnic University, Hong Kong, China\\
$^3$Key Lab of Intelligent Information Processing of Chinese Academy of Sciences (CAS), \\
Institute of Computing Technology, CAS, Beijing 100190, China \\
$^4$CAS Center for Excellence in Brain Science and Intelligence Technology \\}
}

\maketitle

\begin{abstract}
    Deep convolutional networks (CNNs) have exhibited their potential in image inpainting for producing plausible results.
    However, in most existing methods, e.g., context encoder, the missing parts are predicted by propagating the surrounding convolutional features through a fully connected layer, which intends to produce semantically plausible but blurry result.
    In this paper, we introduce a special shift-connection layer to the U-Net architecture, namely Shift-Net, for filling in missing regions of any shape with sharp structures and fine-detailed textures.
    To this end, the encoder feature of the known region is shifted to serve as an estimation of the missing parts.
    A guidance loss is introduced on decoder feature to minimize the distance between the decoder feature after fully connected layer and the ground-truth encoder feature of the missing parts.
    With such constraint, the decoder feature in missing region can be used to guide the shift of encoder feature in known region.
    An end-to-end learning algorithm is further developed to train the Shift-Net.
    Experiments on the Paris StreetView and Places datasets demonstrate the efficiency and effectiveness of our Shift-Net in producing sharper, fine-detailed, and visually plausible results.
    The codes and pre-trained models are available at \url{https://github.com/Zhaoyi-Yan/Shift-Net}.
\keywords{Inpainting, feature rearrangement, deep learning}
\end{abstract}

\section{Introduction}\label{section1}
Image inpainting is the process of filling in missing regions with plausible hypothesis, and can be used in many real world applications such as removing distracting objects, repairing corrupted or damaged parts, and completing occluded regions.
For example, when taking a photo, rare is the case that you are satisfied with what you get directly.
Distracting scene elements, such as irrelevant people or disturbing objects, generally are inevitable but unwanted by the users.
In these cases, image inpainting can serve as a remedy to remove these elements and fill in with plausible content.

%%
%% FIGURE ONE!
%%
\begin{figure}[t]
  \center
\setlength\tabcolsep{0.5pt}
\begin{tabular}{cccc}
    \includegraphics[height=0.24\textwidth]{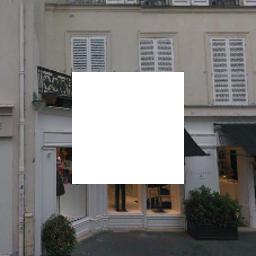} &
    \includegraphics[height=0.24\textwidth]{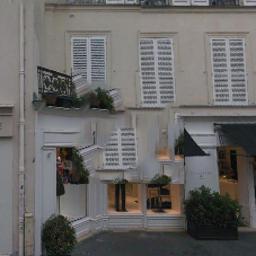} &
    \includegraphics[height=0.24\textwidth]{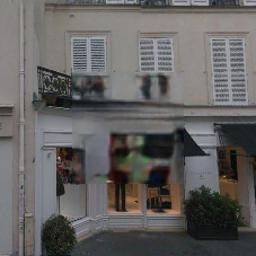} &
    \includegraphics[height=0.24\textwidth]{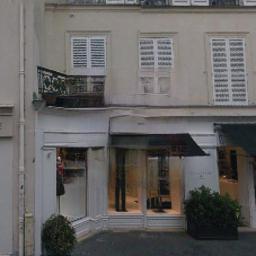} \\
(a) & (b) & (c) & (d) \\
\end{tabular}
\vspace{-1em}
 \caption{Qualitative comparison of inpainting methods. Given (a) an image with a missing region,
 we present the inpainting results by (b) Content-Aware Fill~\cite{Content-Aware-Fill},
 (c) {context encoder}~\cite{pathak2016context}, and (d) our Shift-Net.}
  \vspace{-5mm}
  \label{fig:teaser}
\end{figure}

Despite decades of studies, image inpainting remains a very challenging problem in computer vision and graphics.
In general, there are two requirements for the image inpainting result: (i) global semantic structure and (ii) fine detailed textures.
Classical exemplar-based inpainting methods, e.g., PatchMatch~\cite{barnes2009patchmatch}, gradually synthesize the content of missing parts by searching similar patches from known region.
Even such methods are promising in filling high-frequency texture details, they fail in capturing the global structure of the image (See Fig.~\ref{fig:teaser}(b)).
In contrast, deep convolutional networks (CNNs) have also been suggested to predict the missing parts conditioned on their surroundings~\cite{pathak2016context,yang2017high}.
Benefited from large scale training data, they can produce semantically plausible inpainting result.
However, the existing CNN-based methods usually complete the missing parts by propagating the surrounding convolutional features through a fully connected layer (i.e., bottleneck), making the inpainting results sometimes lack of fine texture details and blurry.
The introduction of adversarial loss is helpful in improving the sharpness of the result, but cannot address this issue essentially (see Fig.~\ref{fig:teaser}(c)).

In this paper, we present a novel CNN, namely Shift-Net, to take into account the advantages of both exemplar-based and CNN-based methods for image inpainting.
Our Shift-Net adopts the U-Net architecture by adding a special shift-connection layer.
In exemplar-based inpainting~\cite{criminisi2003object}, the patch-based replication and filling process are iteratively performed to grow the texture and structure from the known region to the missing parts.
And the patch processing order plays a key role in yielding plausible inpainting result~\cite{le2011examplar, xu2010image}.
We note that CNN is effective in predicting the image structure and semantics of the missing parts.
Guided by the salient structure produced by CNN, the filling process in our Shift-Net can be finished concurrently by introducing a shift-connection layer to connect the encoder feature of known region and the decoder feature of missing parts.
Thus, our Shift-Net inherits the advantages of exemplar-based and CNN-based methods, and can produce inpainting result with both plausible semantics and fine detailed textures (See Fig.~\ref{fig:teaser}(d)).

Guidance loss, reconstruction loss, and adversarial learning are incorporated to guide the shift operation and to learn the model parameters of Shift-Net.
To ensure that the decoder feature can serve as a good guidance, a guidance loss is introduced to enforce the decoder feature be close to the ground-truth encoder feature.
Moreover, $\ell_1$ and adversarial losses are also considered to reconstruct the missing parts and restore more detailed textures.
By minimizing the model objective, our Shift-Net can be end-to-end learned with a training set.
Experiments are conducted on the Paris StreetView dataset~\cite{doersch2012makes}, the Places dataset~\cite{zhou2017places}, and real world images.
The results show that our Shift-Net can handle missing regions with any shape, and is effective in producing sharper, fine-detailed, and visually plausible results (See Fig.~\ref{fig:teaser}(d)).

Besides, Yang \etal.~\cite{yang2017high} also suggest a multi-scale neural patch synthesis (MNPS) approach to incorporating CNN-based with exemplar-based methods.
Their method includes two stages, where an encoder-decoder network is used to generate an initial estimation in the first stage.
By considering both global content and texture losses, a joint optimization model on VGG-19~\cite{simonyan2014very} is minimized to generate the fine-detailed result in the second stage.
Even Yang \etal.~\cite{yang2017high} yields encouraging result, it is very time-consuming and takes about $40,000$ millisecond (ms) to process an image with size of $256 \times 256$.
In contrast, our Shift-Net can achieve comparable or better results (See Fig.~\ref{fig:Paris} and Fig.~\ref{fig:Places} for several examples) and only takes about $80$ ms.
Taking both effectiveness and efficiency into account, our Shift-Net can provide a favorable solution to combine exemplar-based and CNN-based inpainting for improving performance.

To sum up, the main contribution of this work is three-fold:
\begin{enumerate}
  \item By introducing the shift-connection layer to U-Net, a novel Shift-Net architecture is developed to efficiently combine CNN-based and exemplar-based inpainting.
  \item The guidance, reconstruction, and adversarial losses are introduced to train our Shift-Net. Even with the deployment of shift operation, all the network parameters can be learned in an end-to-end manner.
  \item Our Shift-Net achieves state-of-the-art results in comparison with~\cite{barnes2009patchmatch,pathak2016context,yang2017high} and performs favorably in generating fine-detailed textures and visually plausible results.

\end{enumerate}

\begin{figure*}[!t]
  \centering
\begin{overpic}[scale=.12]{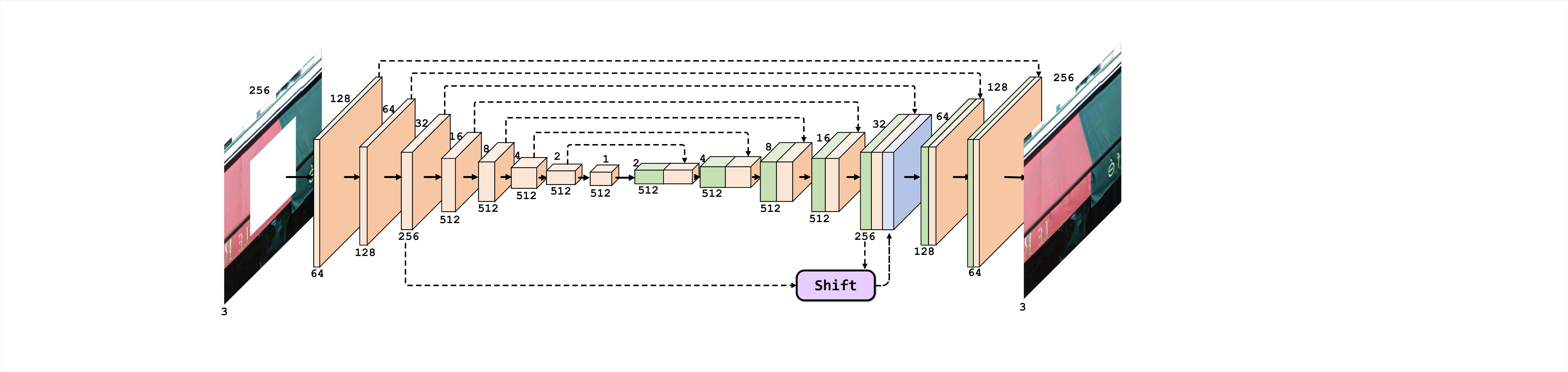} %\transparent{0.4} grid,tics=10

   \put(78,18){\scriptsize {${\Phi_{l}(I)}$}}
   \put(212,22){\scriptsize {${\Phi_{L-l}(I)}$}}
   \put(256,8){\scriptsize {${\Phi_{L-l}^{\text{\emph{shift}}}(I)}$}}

\end{overpic}
\vspace{-1em}
   \caption{The architecture of our model. We add the shift-connection layer at the resolution of $32\times32$.}
   \label{fig:ShiftNetwork}
   \vspace{-1em}
\end{figure*}

\section{Related Work}\label{section2}
In this section, we briefly review the work on each of the three sub-fields, i.e., exemplar-based inpainting, CNN-based inpainting, and style transfer, and specially focus on those relevant to this work.

\vspace{-1em}
\subsection{Exemplar-based inpainting}
In exemplar-based inpainting~\cite{barnes2009patchmatch,barnes2010generalized,criminisi2003object,drori2003fragment,
efros1999texture,jia2003image,jia2004inference,komodakis2006image,komodakis2007image,le2011examplar,pritch2009shift,
simakov2008summarizing,sun2005image,wexler2004space,wexler2007space,xu2010image}, the completion is conducted from the exterior to the interior of the missing part by searching and copying best matching patches from the known region.
For fast patch search, Barnes \etal.~suggest a PatchMatch algorithm~\cite{barnes2009patchmatch} to exploit the image coherency, and generalize it for finding k-nearest neighbors~\cite{barnes2010generalized}.
Generally, exemplar-based inpainting is superior in synthesizing textures, but is not well suited for preserving edges and structures.
For better recovery of image structure, several patch priority measures have been proposed to fill in structural patches first~\cite{criminisi2003object,le2011examplar, xu2010image}.
Global image coherence has also been introduced to the Markov random field (MRF) framework for improving visual quality~\cite{komodakis2006image, pritch2009shift, wexler2004space}.
However, these methods only work well on images with simple structures, and may fail in handling images with complex objects and scenes.
Besides, in most exemplar-based inpainting methods~\cite{komodakis2006image, komodakis2007image, pritch2009shift}, the missing part is recovered as the shift representation of the known region in pixel/region level, which also motivates our shift operation on convolution feature representation.
\vspace{-1em}

\subsection{CNN-based inpainting}
Recently, deep CNNs have achieved great success in image inpainting.
Originally, CNN-based inpainting is confined to small and thin masks~\cite{kohler2014mask, ren2015shepard, xie2012image}.
Phatak \etal.~\cite{pathak2016context} present an encoder-decoder~(i.e., context encoder) network to predict the missing parts, where an adversarial loss is adopted in training to improve the visual quality of the inpainted image.
Even context encoder is effective in capturing image semantics and global structure, it completes the input image with only one forward-pass and performs poorly in generating fine-detailed textures.
%
% Beacuse pix2pix actually aims not at inpainting, and they haven't release their model for inpainting(Just image translation models)
% Besides, I cannot get the similar results from their code(Have used their proposed training settings for inpainting).
% All the pix2pix models collapse and get very bad results.
% So I think maybe it is better not to explictly add this to avoid comparisons with them.
%
%{\color{red}{\sout{Other network architecture, e.g., U-Net~\cite{ronneberger2015u}, has been employed to image translation~\cite{zhu2017unpaired} and can also be adopted in image inpainting~\cite{isola2016image}}}}.
%
Semantic image inpainting is introduced to fill in the missing part conditioned on the known region for images from a specific semantic class~\cite{yeh2017semantic}.
In order to obtain globally consistent result with locally realistic details, global and local discriminators have been proposed in image inpainting~\cite{IizukaSIGGRAPH2017} and face completion~\cite{li2017generative}.
For better recovery of fine details, MNPS is presented to combine exemplar-based and CNN-based inpainting~\cite{yang2017high}.

\vspace{-1em}
\subsection{Style transfer}
Image inpainting can be treated as an extension of style transfer, where both the content and style (texture) of missing part are estimated and transferred from the known region.
In the recent few years, style transfer~\cite{chen2016fast, dumoulin2016learned, gatys2015neural, gatys2016controlling, huang2017arbitrary, johnson2016perceptual, li2016combining, luan2017deep, ulyanov2016texture} has been an active research topic.
Gatys \etal.~\cite{gatys2015neural} show that one can transfer style and texture of the style image to the content image by solving an optimization objective defined on an existing CNN.
Instead of the Gram matrix, Li \etal.~\cite{li2016combining} apply the MRF regularizer to style transfer to suppress distortions and smears.
In~\cite{chen2016fast}, local matching is performed on the convolution layer of the pre-trained network to combine content and style, and an inverse network is then deployed to generate the image from feature representation.

\section{Method}\label{section3}
Given an input image $I$, image inpainting aims to restore the ground-truth image $I^{gt}$ by filling in the missing part.
To this end, we adopt U-Net~\cite{ronneberger2015u} as the baseline network.
By incorporating with guidance loss and shift operation, we develop a novel Shift-Net for better recovery of semantic structure and fine-detailed textures.
In the following, we first introduce the guidance loss and Shift-Net, and then describe the model objective and learning algorithm.

\vspace{-1em}
\subsection{Guidance loss on decoder feature}\label{section3.1}
The U-Net consists of an encoder and a symmetric decoder, where skip connection is introduced to concatenate the features from each layer of encoder and those of the corresponding layer of decoder.
Such skip connection makes it convenient to utilize the information before and after bottleneck, which is valuable for image inpainting and other low level vision tasks in capturing localized visual details~\cite{isola2016image, zhu2017unpaired}.
The architecture of the U-Net adopted in this work is shown in Fig.~\ref{fig:ShiftNetwork}. Please refer to the supplementary material for more details on network parameters.

%For image inpainting, the U-Net possesses more characteristics which are useful in improving completion result.
%
Let $\Omega$ be the missing region and $\overline{\Omega}$ be the known region.
Given a U-Net of $L$ layers, $\Phi_{l}(I)$ is used to denote the encoder feature of the $l$-th layer, and $\Phi_{L-l}(I)$ the decoder feature of the $(L-l)$-th layer.
For the end of recovering $I^{gt}$, we expect that $\Phi_{l}(I)$ and $\Phi_{L-l}(I)$ convey almost all the information in $\Phi_{l}(I^{gt})$.
For any location $\mathbf{y} \in \Omega$, we have $\left( \Phi_{l}(I) \right)_{\mathbf{y}} \approx 0$.
Thus, $\left( \Phi_{L-l}(I) \right)_{\mathbf{y}}$ should convey equivalent information of $\left( \Phi_{l}(I^{gt}) \right)_{\mathbf{y}}$.

In this work, we suggest to explicitly model the relationship between  $\left( \Phi_{L-l}(I) \right)_{\mathbf{y}}$ and $\left( \Phi_{l}(I^{gt}) \right)_{\mathbf{y}}$ by introducing the following guidance loss,
\begin{equation}\label{loss_guidance}
\small
{\cal L}_g = \sum_{\mathbf{y} \in \Omega} \left\| \left( \Phi_{L-l}(I) \right)_{\mathbf{y}} - \left( \Phi_{l}(I^{gt}) \right)_{\mathbf{y}} \right\|_2^2.
\end{equation}
We note that $\left( \Phi_{l}(I) \right)_{\mathbf{x}} \approx \left( \Phi_{l}(I^{gt}) \right)_{\mathbf{x}}$ for any $\mathbf{x} \in \overline{\Omega}$.
Thus the guidance loss is only defined on $\mathbf{y} \in {\Omega}$ to make $\left( \Phi_{L-l}(I) \right)_{\mathbf{y}} \approx \left( \Phi_{l}(I^{gt}) \right)_{\mathbf{y}}$.
By concatenating $\Phi_{l}(I)$ and $\Phi_{L-l}(I)$, all information in $\Phi_{l}(I^{gt})$ can be approximately obtained.

\begin{figure}[t]
  \center
\setlength\tabcolsep{0.5pt}
\begin{tabular}{cccc}
    \includegraphics[height=0.24\textwidth]{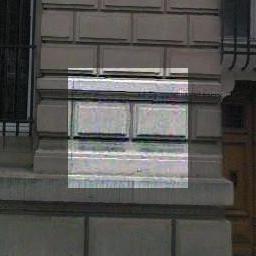} &
    \includegraphics[height=0.24\textwidth]{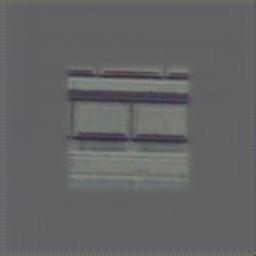} &
    \includegraphics[height=0.24\textwidth]{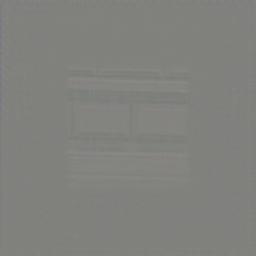} &
    \includegraphics[height=0.24\textwidth]{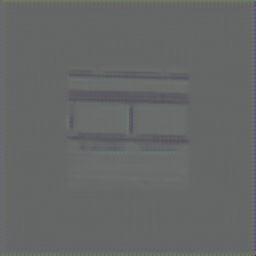} \\
 (a) & (b) & (c) & (d) \\
\end{tabular}
  \vspace{-1mm}
\caption{Visualization of features learned by our model. Given (a) an input image, (b) is the visualization of $\left( \Phi_{l}(I^{gt})\right)_{\mathbf{y}}$ (i.e., $H^{gt}$),
(c) shows the result of $\left( \Phi_{L-l}(I) \right)_{\mathbf{y}}$ (i.e., $H^{de}$) and
(d) demonstrates the effect of $\left( \Phi_{L-l}^{\text{\emph{shift}}}(I) \right)_{\mathbf{y}}$.}
  \vspace{-5mm}
  \label{fig:Visualization}
\end{figure}

Experiment on deep feature visualization is further conducted to illustrate the relation between $\left( \Phi_{L-l}(I) \right)_{\mathbf{y}}$ and $\left( \Phi_{l}(I^{gt}) \right)_{\mathbf{y}}$.
For visualizing $\{ \left( \Phi_{l}(I^{gt}) \right)_{\mathbf{y}} | {\mathbf{y}} \in \Omega \}$, we adopt the method~\cite{mahendran2015understanding} by solving an optimization problem
\begin{equation}\label{vis_gt}
\small
H^{gt} = \arg \min_{H} \sum_{{\mathbf{y}} \in \Omega} \left\| \left( \Phi_{l}(H) \right)_{\mathbf{y}} - \left( \Phi_{l}(I^{gt}) \right)_{\mathbf{y}}\right\|_2^2.
\end{equation}
Analogously, $\{ \left( \Phi_{L-l}(I) \right)_{\mathbf{y}} | {\mathbf{y}} \in \Omega \}$ is visualized by
\begin{equation}\label{vis_gt}
\small
H^{de} = \arg \min_{H} \sum_{{\mathbf{y}} \in \Omega} \left\| \left( \Phi_{l}(H) \right)_{\mathbf{y}} - \left( \Phi_{L-l}(I) \right)_{\mathbf{y}} \right\|_2^2.
\end{equation}
Figs.~\ref{fig:Visualization}(b)(c) show the visualization results of $H^{gt}$ and $H^{de}$.
With the introduction of guidance loss, obviously $H^{de}$ can serve as a reasonable estimation of $H^{gt}$, and U-Net works well in recovering image semantics and structures.
However, in compared with $H^{gt}$ and $I^{gt}$, the result $H^{de}$ is blurry, which is consistent with the poor performance of CNN-based inpainting in recovering fine textures~\cite{yang2017high}.
Finally, we note that the guidance loss is helpful in constructing an explicit relation between $\left( \Phi_{L-l}(I) \right)_{\mathbf{y}}$ and $\left( \Phi_{l}(I^{gt}) \right)_{\mathbf{y}}$.
In the next section, we will explain how to utilize such property for better estimation to $\left( \Phi_{l}(I^{gt}) \right)_{\mathbf{y}}$ and enhancing inpainting result.

%%%%%%%%%%%%%%%%%%%%%%%%%%%%%%%%%%%%%%%%%%%%%%%%%%%%%%%%%%%%
%%%%%%%%%%%%%%%%%% Figure Visualization %%%%%%%%%%%%%%%%%
%%%%%%%%%%%%%%%%%%%%%%%%%%%%%%%%%%%%%%%%%%%%%%%%%%%%%%%%%%%%

\vspace{-1em}
\subsection{Shift operation and Shift-Net}\label{section3.2} %% Need define
In exemplar-based inpainting, it is generally assumed that the missing part is the spatial rearrangement of the pixels/patches in the known region.
For each pixel/patch localized at $\mathbf{y}$ in missing part, exemplar-based inpainting explicitly or implicitly find a shift vector $\mathbf{u}_{\mathbf{y}}$, and recover $(I)_{\mathbf{y}}$ with $(I)_{\mathbf{y} + \mathbf{u}_{\mathbf{y}} }$, where ${\mathbf{y}+\mathbf{u}_{\mathbf{y}} } \in \overline{\Omega}$ is in the known region.
The pixel value $(I)_{\mathbf{y}}$ is unknown before inpainting.
Thus, the shift vectors usually are obtained progressively from the exterior to the interior of the missing part, or by solving a MRF model by considering global image coherence.
However, these methods may fail in recovering complex image semantics and structures.

We introduce a special shift-connection layer in U-Net, which takes $\Phi_{l}(I)$ and $\Phi_{L-l}(I)$ to obtain an updated estimation on $\Phi_{l}(I^{gt})$.

%As illustrated in Fig.~\ref{fig:ShiftConnection}, we introduce a special shift-connection layer in U-Net, which takes $\Phi_{l}(I)$ and $\Phi_{L-l}(I)$ to obtain an updated estimation on $\Phi_{l}(I^{gt})$.
%
For each $\left( \Phi_{L-l}(I) \right)_{\mathbf{y}}$ with $\mathbf{y} \in \Omega$, its nearest neighbor searching in $\left( \Phi_{l}(I) \right)_{\mathbf{x}}$ ($\mathbf{x} \in \overline{\Omega}$) can be independently obtained by,
\begin{equation}\label{eqn:nn}
\mathbf{x}^*(\mathbf{y}) = \arg \max_{\mathbf{x} \in \overline{\Omega}}
\frac{\left \langle \left( \Phi_{L-l}(I) \right)_{\mathbf{y}}, \left( \Phi_{l}(I) \right)_{\mathbf{x}} \right \rangle}
{\|\left( \Phi_{L-l}(I) \right)_{\mathbf{y}}\|_2  \|\left( \Phi_{l}(I) \right)_{\mathbf{x}}\|_2},
\end{equation}
and the shift vector is defined as $\mathbf{u}_{\mathbf{y}} = \mathbf{x}^*(\mathbf{y}) - \mathbf{y}$.
Similar to~\cite{li2016combining}, the nearest neighbor searching can be computed as a convolutional layer.
Then, we update the estimation of $\left( \Phi_{l}(I^{gt}) \right)_{\mathbf{y}}$ as the spatial rearrangement of the encoder feature $\left( \Phi_{l}(I) \right)_{\mathbf{x}}$,
\begin{equation}\label{eqn:shift}
\left( \Phi_{L-l}^{\text{\emph{shift}}}(I) \right)_{\mathbf{y}} = \left( \Phi_{l}(I) \right)_{\mathbf{y} + \mathbf{u}_{\mathbf{y}}}.
\end{equation}
See Fig.~\ref{fig:Visualization}(d) for visualization.
Finally, as shown in Fig.~\ref{fig:ShiftNetwork}, the convolution features $\Phi_{L-l}(I)$, $\Phi_{l}(I)$ and $\Phi_{L-l}^{\text{\emph{shift}}}(I)$ are concatenated and taken as inputs to the $(L-l+1)$-th layer, resulting in our Shift-Net.

%%%%%%%%%%%%%%%%%% Figure Shift-Connection %%%%%%%%%%%%%%%%%%%%%%%%%%%
%%%%%%%%%%%%%%%%%%%%%%%%%%%%%%%%%%%%%%%%%%%%%%%%%%%%%%%%%%%%%%%%%%%%%%
%\begin{figure}[t]
%\vspace{-.4em}
%\begin{center}
%\includegraphics[scale=.10]{Shift-Connection.pdf}
%\end{center}
%\vspace{-2em}
%   \caption{Illustration of the shift operation.}
%   \label{fig:ShiftConnection}
%   \vspace{-2em}
%\end{figure}

The shift operation is different with exemplar-based inpainting from several aspects.
(i) While exemplar-based inpainting is operated on pixels/patches, shift operation is performed on deep encoder feature domain which is end-to-end learned from training data.
(ii) In exemplar-based inpainting, the shift vectors are obtained either by solving an optimization problem or in particular order. As for shift operation, with the guidance of $\Phi_{L-l}(I)$, all the shift vectors can be computed in parallel.
(iii) For exemplar-based inpainting, both patch processing orders and global image coherence are not sufficient for preserving complex structures and semantics. In contrast, in shift operation $\Phi_{L-l}(I)$ is learned from large scale data and is more powerful in capturing global semantics.
(iv) In exemplar-based inpainting, after obtaining the shift vectors, the completion result can be directly obtained as the shift representation of the known region. As for shift operation, we take the shift representation $\Phi_{L-l}^{\text{\emph{shift}}}(I)$ together with $\Phi_{L-l}(I)$ and $\Phi_{l}(I)$ as inputs to $(L-l+1)$-th layer of U-Net, and adopt a data-driven manner to learn an appropriate model for image inpainting.
Moreover, even with the introduction of shift-connection layer, all the model parameters in our Shift-Net can be end-to-end learned from training data.
Thus, our Shift-Net naturally inherits the advantages of exemplar-based and CNN-based inpainting.
%\vspace{-1em}

\vspace{-1em}
\subsection{Model objective and learning}\label{section3.3}

\subsubsection{Objective.}\label{section3.3.1}

Denote by $\Phi(I; \mathbf{W})$ the output of our Shift-Net, where $\mathbf{W}$ is the model parameters to be learned.
Besides the guidance loss, the $\ell_1$ loss and the adversarial loss are also included to train our Shift-Net.
The $\ell_1$ loss is defined as,
\begin{equation}\label{eqn:l2loss}
{\cal L}_{\ell_{1}} = \|\Phi(I; \mathbf{W}) - I^{gt}\|_1,
\end{equation}
which is suggested to constrain that the inpainting result should approximate the ground-truth image.

Recently, adversarial learning has been adopted in many low level vision~\cite{ledig2016photo} and image generation tasks\cite{isola2016image, radford2015unsupervised}, and exhibits its superiority in restoring high-frequency details and photo-realistic textures.
As for image inpainting, we use $p_{data}(I^{gt})$ to denote the distribution of ground-truth images, and $p_{miss}(I)$ to denote the distribution of input image.
The adversarial loss is then defined as,
\begin{align}\label{eqn:ganloss}
{\cal L}_{adv} \!&\! =  \min_{\mathbf{W}} \max_{D}  \mathbb{E}_{I^{gt} \sim p_{data}({I^{gt}})} [\log D({I^{gt}})]  \\
 \!&\! + \mathbb{E}_{I \sim p_{miss}({I})} [\log ( 1 - D(\Phi(I; \mathbf{W})) )],
\end{align}
where $D(\cdot)$ denotes the discriminator to predict the probability that an image is from the distribution $p_{data}(I^{gt})$.

Taking guidance, $\ell_1$, and adversarial losses into account, the overall objective of our Shift-Net is defined as,
\begin{equation}\label{eqn:objective}
{\cal L} = {\cal L}_{\ell_{1}} + \lambda_{g} {\cal L}_g + \lambda_{adv} {\cal L}_{adv},
\end{equation}
where $\lambda_{g}$  and $\lambda_{adv}$ are the tradeoff parameters for the guidance and adversarial losses, respectively.

\subsubsection{Learning.}\label{section3.3.2}

Given a training set $\{ (I, I^{gt}) \}$, the Shift-Net is trained by minimizing the objective in Eqn.~(\ref{eqn:objective}) via back-propagation.
We note that the Shift-Net and the discriminator are trained in an adversarial manner.
The Shift-Net $\Phi(I; \mathbf{W})$ is updated by minimizing the adversarial loss ${\cal L}_{adv}$, while the discriminator $D$ is updated by maximizing ${\cal L}_{adv}$.

Due to the introduction of shift-connection layer, we should modify the computation of the gradient w.r.t. the $l$-th layer of feature $F_l = \Phi_{l}(I)$.
To avoid confusion, we use $F_l^{skip}$ to denote the feature $F_l$ after skip connection, and of course we have $F_l^{skip} = F_l$.
According to Eqn.~(\ref{eqn:shift}), the relation between ${\Phi_{L-l}^{\text{\emph{shift}}}(I)}$ and $\Phi_{l}(I)$ can be written as,
\begin{equation}\label{BPEqn1}
{\Phi_{L-l}^{\text{\emph{shift}}}(I)} = \mathbf{P} \Phi_{l}(I),
\end{equation}
where $\mathbf{P}$ denotes the shift matrix of $\{0, 1\}$, and there is only one element of 1 in each row of $\mathbf{P}$.
Thus, the gradient with respect to $\Phi_{l}(I)$ consists of three terms,
\begin{equation}\label{eqn:gradient}
%\small
\frac{\partial {\cal L}}{\partial F_l} \!=\! \frac{\partial {\cal L}}{\partial F_l^{skip}}
\!+\! \frac{\partial {\cal L}}{\partial F_{l+1}} \frac{\partial F_{l+1}} {\partial F_l}
\!+\! \mathbf{P}^T \frac{\partial {\cal L}}{\partial \Phi_{L-l}^{\text{\emph{shift}}}(I)},
\end{equation}
where the computation of the first two terms are the same with U-Net, and the gradient with respect to ${ \Phi_{L-l}^{\text{\emph{shift}}}(I)}$ can also be directly computed.
Thus, our Shift-Net can also be end-to-end trained to learn the model parameters $\mathbf{W}$.

\section{Experiments}\label{section4}
We evaluate our method on two datasets: Paris StreetView~\cite{doersch2012makes} and six scenes from Places365-Standard dataset~\cite{zhou2017places}.
The Paris StreetView contains 14,900 training images and 100 test images.
There are 1.6 million training images from 365 scene categories in the Places365-Standard.
The scene categories selected from Places365-Standard are \emph{butte}, \emph{canyon}, \emph{field}, \emph{synagogue}, \emph{tundra} and \emph{valley}.
Each category has 5,000 training images, 900 test images and 100 validation images.
Our model is learned using the training set and tested on the validation set.
For both Paris StreetView and Places, we resize each training image to let its minimal length/width be 350, and randomly crop a subimage of size $256\times256$ as input to our model.
Moreover, our method is also tested on real world images for removing objects and distractors.
Our Shift-Net is optimized using the Adam algorithm~\cite{kingma2015adam} with a learning rate of $2 \times {10^{ - 4}}$ and ${\beta _1} = 0.5$.
The batch size is $1$ and the training is stopped after $30$ epochs.
Data augmentation such as flipping is also adopted during training.
The tradeoff parameters are set as $\lambda_{g} = 0.01$ and $\lambda_{adv} = 0.002$.
It takes about one day to train our Shift-Net on an Nvidia Titan X Pascal GPU.

\vspace{-1em}

%%% Figure.Comparisons on Paris
\begin{figure*}[!t]
  \center
\setlength\tabcolsep{1.5pt}
\begin{tabular}{ccccc}
  \includegraphics[width=.18\textwidth]{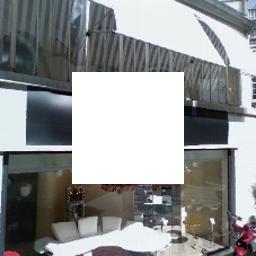} &
  \includegraphics[width=.18\textwidth]{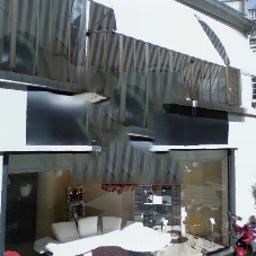} &
  \includegraphics[width=.18\textwidth]{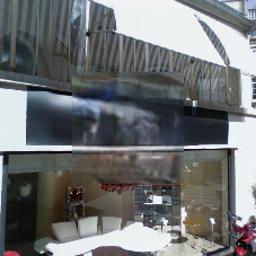} &
  \includegraphics[width=.18\textwidth]{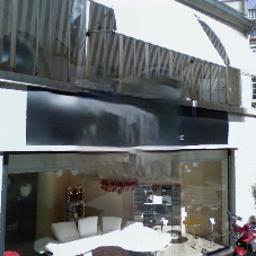} &
  \includegraphics[width=.18\textwidth]{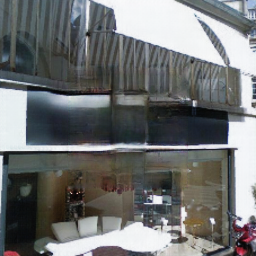}\\

  \includegraphics[width=.18\textwidth]{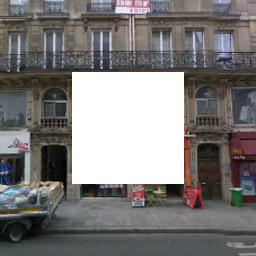} &
  \includegraphics[width=.18\textwidth]{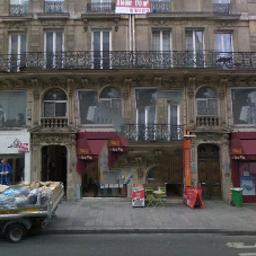} &
  \includegraphics[width=.18\textwidth]{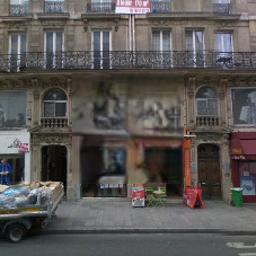} &
  \includegraphics[width=.18\textwidth]{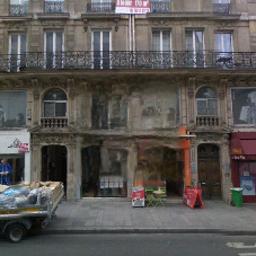} &
  \includegraphics[width=.18\textwidth]{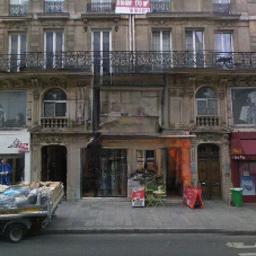}\\

  \includegraphics[width=.18\textwidth]{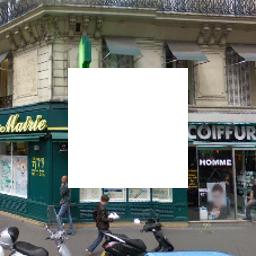} &
  \includegraphics[width=.18\textwidth]{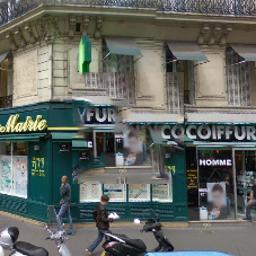}  &
  \includegraphics[width=.18\textwidth]{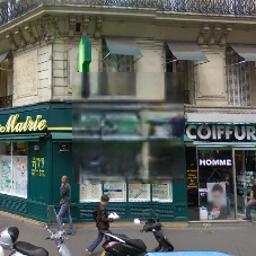} &
  \includegraphics[width=.18\textwidth]{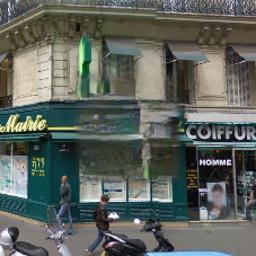} &
  \includegraphics[width=.18\textwidth]{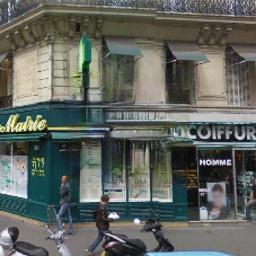}\\

(a)  & (b) & (c)  & (d) & (e)\\
\end{tabular}
\vspace{-.5em}
\caption{Qualitative comparisons on the Paris StreetView dataset. From the left to the right are:
(a) input, (b) Content-Aware Fill~\cite{Content-Aware-Fill}, (c) context encoder~\cite{pathak2016context}, (d) MNPS~\cite{yang2017high} and (e) Ours. All images are scaled to $256\times 256$.}
\label{fig:Paris}
\vspace{-1em}
\end{figure*}

\begin{figure}[!t]
\setlength\tabcolsep{1.5pt}
\centering
\small
\begin{tabular}{ccccc}
  \includegraphics[width=.18\linewidth]{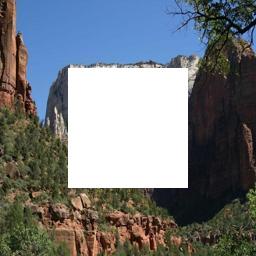} &
  \includegraphics[width=.18\linewidth]{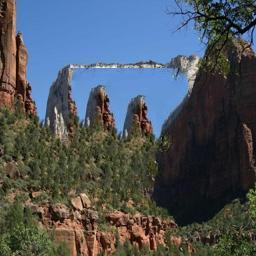} &
  \includegraphics[width=.18\linewidth]{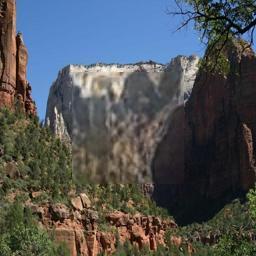} &
  \includegraphics[width=.18\linewidth]{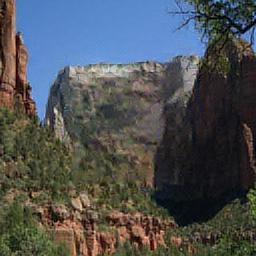} &
  \includegraphics[width=.18\linewidth]{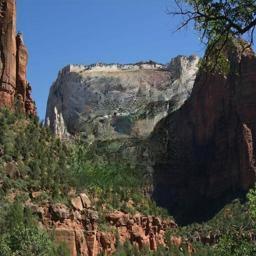}\\

  \includegraphics[width=.18\linewidth]{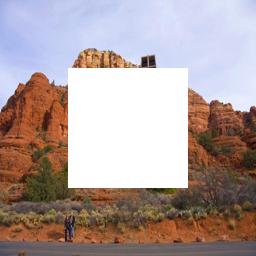} &
  \includegraphics[width=.18\linewidth]{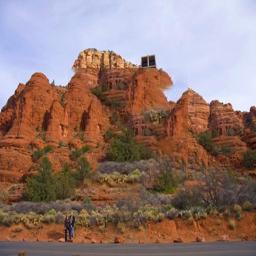} &
  \includegraphics[width=.18\linewidth]{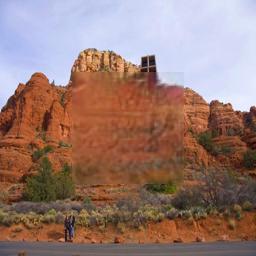} &
  \includegraphics[width=.18\linewidth]{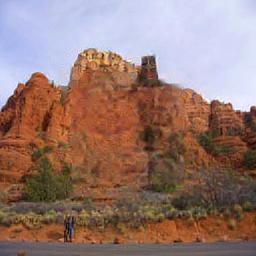} &
  \includegraphics[width=.18\linewidth]{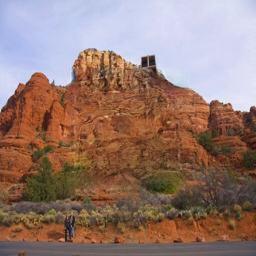}\\

  \includegraphics[width=.18\linewidth]{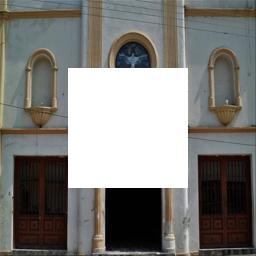} &
  \includegraphics[width=.18\linewidth]{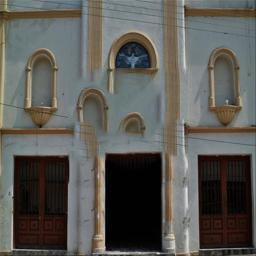} &
  \includegraphics[width=.18\linewidth]{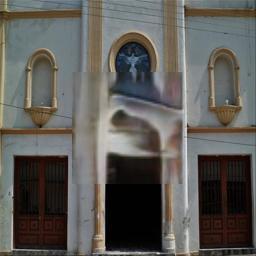} &
  \includegraphics[width=.18\linewidth]{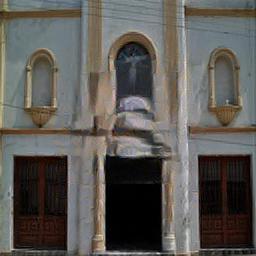} &
  \includegraphics[width=.18\linewidth]{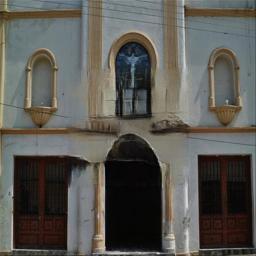}\\
(a)  & (b) & (c)  & (d) & (e)\\
\end{tabular}
\vspace{-.5em}
\caption{Qualitative comparisons on the Places. From the left to the right are:
(a) input, (b) Content-Aware Fill~\cite{Content-Aware-Fill}, (c) context encoder~\cite{pathak2016context}, (d) MNPS~\cite{yang2017high} and (e) Ours. All images are scaled to $256\times 256$.}
\label{fig:Places}
\end{figure}

\subsection{Comparisons with state-of-the-arts}\label{section4.1}
We compare our results with Photoshop Content-Aware Fill~\cite{Content-Aware-Fill} based on~\cite{barnes2009patchmatch}, context encoder~\cite{pathak2016context}, and MNPS~\cite{yang2017high}.
As context encoder only accepts $128 \times 128$ images, we upsample the results to $256 \times 256$.
For MNPS~\cite{yang2017high}, we set the pyramid level be 2 to get the resolution of $256 \times 256$.

\noindent\textbf{Evaluation on Paris StreetView and Places.}
Fig.~\ref{fig:Paris} shows the comparisons of our method with the three state-of-the-art approaches on Paris StreetView.
Content-Aware Fill~\cite{Content-Aware-Fill} is effective in recovering low level textures, but performs slightly worse in handling occlusions with complex structures.
Context encoder~\cite{pathak2016context} is effective in semantic inpainting, but the results seem blurry and detail-missing due to the effect of bottleneck.
MNPS~\cite{yang2017high} adopts a multi-stage scheme to combine CNN and examplar-based inpainting, and generally works better than Content-Aware Fill~\cite{Content-Aware-Fill} and context encoder~\cite{pathak2016context}.
However, the multi-scales in MNPS~\cite{yang2017high} are not jointly trained, where some adverse effects produced in the first stage may not be eliminated by the subsequent stages.
In comparison to the competing methods, our Shift-Net combines CNN and examplar-based inpainting in an end-to-end manner, and generally is able to generate visual-pleasing results.
Moreover, we also note that our Shift-Net is much more efficient than MNPS~\cite{yang2017high}.
Our method consumes only about $80$ ms for a $256 \times 256$ image, which is about 500$\times$ faster than MNPS~\cite{yang2017high} (about $40$ seconds).
In addition, we also evaluate our method on the Places dataset (see Fig.~\ref{fig:Places}).
Again our Shift-Net performs favorably in generating fine-detailed, semantically plausible, and realistic images.

\noindent\textbf{Quantitative evaluation.}
We also compare our model quantitatively with the competing methods on the Paris StreetView dataset.
Table \ref{table:paris} lists the PSNR, SSIM and mean \(\ell_2\) loss of different methods.
Our Shift-Net achieves the best numerical performance.
We attribute it to the combination of CNN-based with examplar-based inpainting as well as the end-to-end training.
In comparison, MNPS~\cite{yang2017high} adopts a two-stage scheme and cannot be jointly trained.

%% PSNR, SSIM : Table of comparisons of methods for each image.

\begin{table}[!t]
 \scriptsize
  \caption{Comparison of PSNR, SSIM and mean \(\ell_2\) loss on Paris StreetView dataset.}
  \vspace{-1.5em}
\begin{center}
\resizebox{.90\textwidth}{!}{%
  \begin{tabular}{ l  c  c c}
    \hline
    Method & PSNR & SSIM & Mean \(\ell_2\) Loss\\ \hline
    Content-Aware Fill~\cite{Content-Aware-Fill} & 23.71 & 0.74 & 0.0617 \\ \hline
    context encoder~\cite{pathak2016context} (\(\ell_2\) + adversarial loss) & 24.16 & 0.87 &  0.0313 \\ \hline
    MNPS~\cite{yang2017high} & 25.98 & 0.89 & 0.0258 \\ \hline
    Ours & \textbf{26.51} & \textbf{0.90} & \textbf{0.0208} \\ \hline
  \end{tabular}}
  \end{center}
  \label{table:paris}
  \vspace{-3em} %**********************************************************************************
\end{table}

\begin{figure}[!t]
  \centering
\setlength\tabcolsep{1.5pt}
\begin{tabular}{ccccc}
  \includegraphics[width=.18\linewidth]{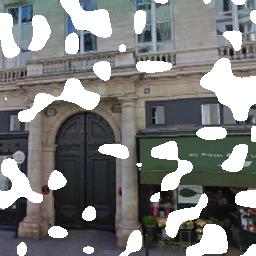}&
  \includegraphics[width=.18\linewidth]{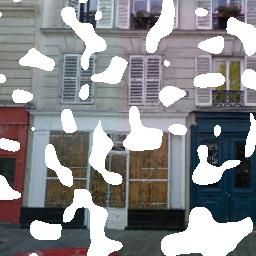}&
  \includegraphics[width=.18\linewidth]{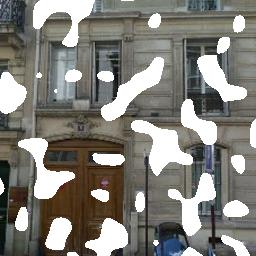}&
  \includegraphics[width=.18\linewidth]{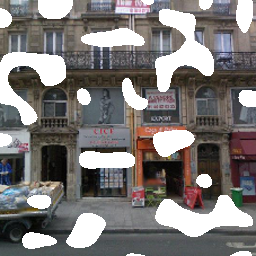}&
  \includegraphics[width=.18\linewidth]{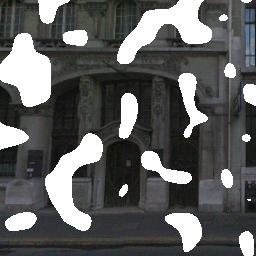}\\

  \includegraphics[width=.18\linewidth]{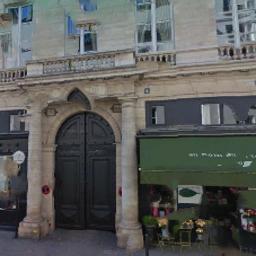}&
  \includegraphics[width=.18\linewidth]{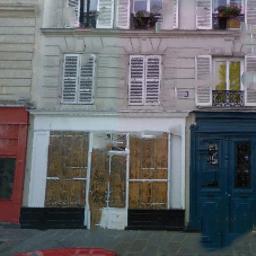}&
  \includegraphics[width=.18\linewidth]{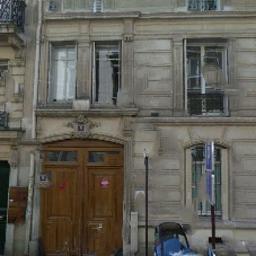}&
  \includegraphics[width=.18\linewidth]{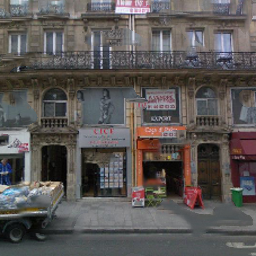}&
  \includegraphics[width=.18\linewidth]{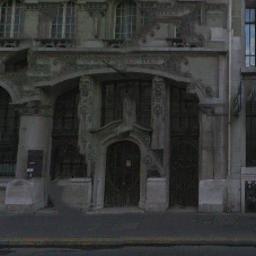}\\

  \includegraphics[width=.18\linewidth]{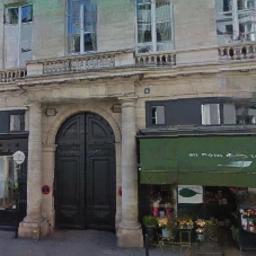}&
  \includegraphics[width=.18\linewidth]{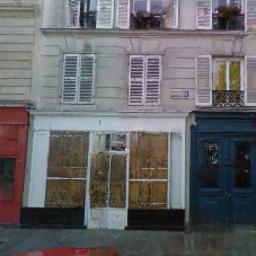}&
  \includegraphics[width=.18\linewidth]{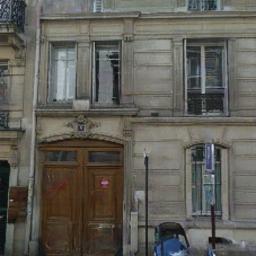}&
  \includegraphics[width=.18\linewidth]{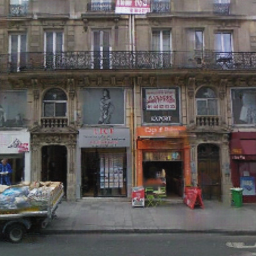}&
  \includegraphics[width=.18\linewidth]{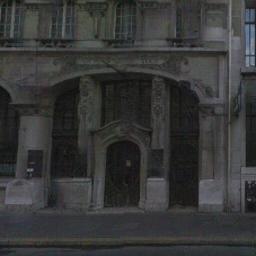}\\
\end{tabular}
\vspace{-1em}
\caption{Random region completion. From top to bottom are: input, Content-Aware Fill~\cite{Content-Aware-Fill}, and Ours.}
\label{fig:Paris_random}
\end{figure}

\noindent\textbf{Random mask completion.}
Our model can also be trained for arbitrary region completion.
Fig.~\ref{fig:Paris_random} shows the results by Content-Aware Fill~\cite{Content-Aware-Fill} and our Shift-Net.
For textured and smooth regions, both Content-Aware Fill~\cite{Content-Aware-Fill} and our Shift-Net perform favorably.
While for structural region, our Shift-Net is more effective in filling the cropped regions with context coherent with global content and structures.
%
%%%%%%%%%%%%%%%%%%%%%%%%%%%%%%%%
%%% Figure.Random mask on Paris
%%%%%%%%%%%%%%%%%%%%%%%%%%%%%%%%

\subsection{Inpainting of real world images}\label{section4.2}

We also evaluate our Shift-Net trained on Paris StreetView for the inpainting of real world images by considering two types of missing regions: (i) central region, (ii) object removal.
From the first row of Fig.~\ref{fig:realImgs}, one can see that our Shift-Net trained with central mask can be generalized to handle real world images.
From the second row of Fig.~\ref{fig:realImgs}, we show the feasibility of using our Shift-Net trained with random mask to remove unwanted objects from the images.

%%%%%%%%%%%%%%%%%%%%%%%%%%%%%%%%%%%%%%%%%%%%%%
%Figure.Comparisons on real images.(two images)
%%%%%%%%%%%%%%%%%%%%%%%%%%%%%%%%%%%%%%%%%%%%%%
\begin{figure}[!t]
%\vspace{-2em}
\setlength\tabcolsep{1.5pt}
\centering
\begin{tabular}{cccc}

\includegraphics[width=.24\linewidth]{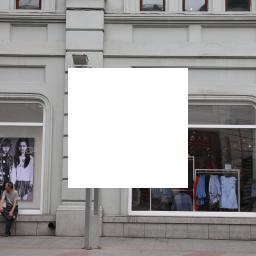}&
\includegraphics[width=.24\linewidth]{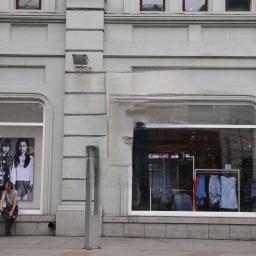}&
\includegraphics[width=.24\linewidth]{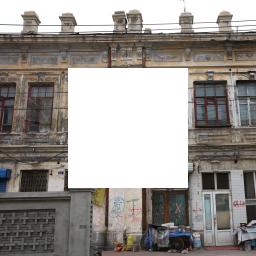}&
\includegraphics[width=.24\linewidth]{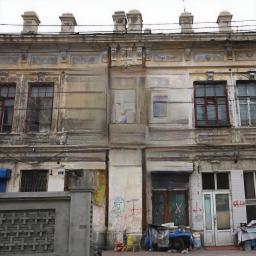}\\
\includegraphics[width=.24\linewidth]{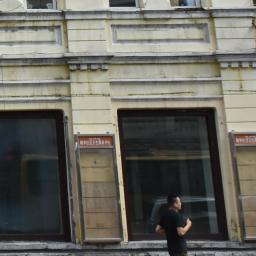}&
\includegraphics[width=.24\linewidth]{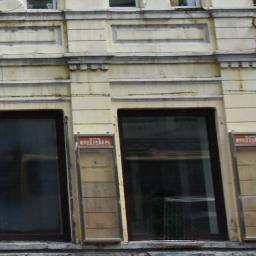}&
\includegraphics[width=.24\linewidth]{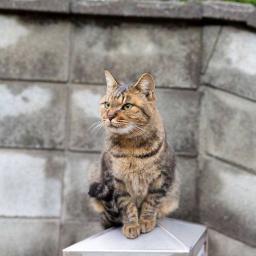}&
\includegraphics[width=.24\linewidth]{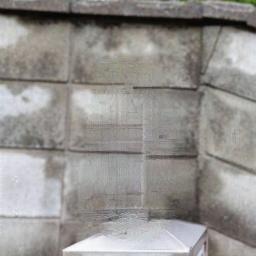}\\

\end{tabular}
\vspace{-1em}
\caption{Results on real images. From the top to bottom are: central region inpainting, and object removal.}
\label{fig:realImgs}
\end{figure}

\section{Ablative Studies}\label{section5}
%%%%%%%%%%%%%%%%%%%%%%%%%%%%%%%%%%%%%%%%%%%%%%%%%%
%% Comparison of guidance loss in different models.
%%%%%%%%%%%%%%%%%%%%%%%%%%%%%%%%%%%%%%%%%%%%%%%%%%
\begin{figure}[!t]
\vspace{-0em}
\setlength\tabcolsep{1.5pt}
\centering
\small
\begin{tabular}{cccc}
\includegraphics[width=.24\linewidth]{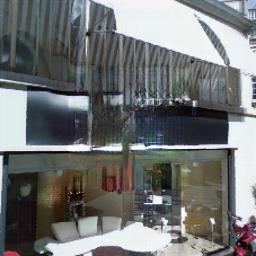} &
\includegraphics[width=.24\linewidth]{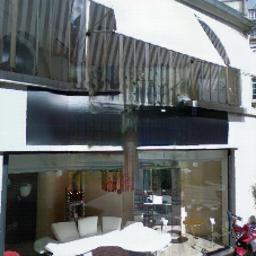} &
\includegraphics[width=.24\linewidth]{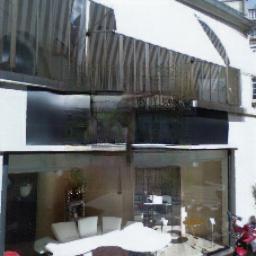} &
\includegraphics[width=.24\linewidth]{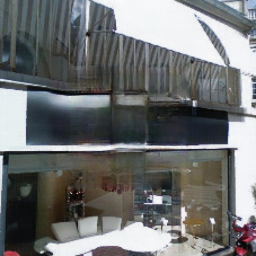} \\
(a) U-Net & (b) U-Net  & (c) Ours  & (d) Ours\\
(w/o ${\cal L}_g$) & (w/ ${\cal L}_g$) & (w/o ${\cal L}_g$) & (w/ ${\cal L}_g$)
\end{tabular}
\vspace{-1em}
\caption{The effect of guidance loss ${\cal L}_g$ in U-Net and our Shift-Net. }
\label{fig:guidanceEffectiveness}
\vspace{-1em}
\end{figure}

%%%%%%%%%%%%%%%%%%%%%%%%%%%%%%%%%%%%%%%%%%%%%%%%%%
%% comparisons of differnet weight of guidance loss.
%%%%%%%%%%%%%%%%%%%%%%%%%%%%%%%%%%%%%%%%%%%%%%%%%%

\begin{figure}[!t]
\setlength\tabcolsep{1.5pt}
\centering
\small
\begin{tabular}{cccc}
\includegraphics[width=.24\linewidth]{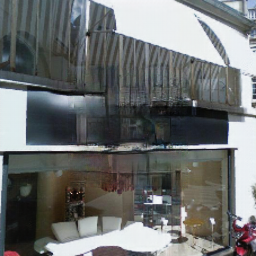} &
\includegraphics[width=.24\linewidth]{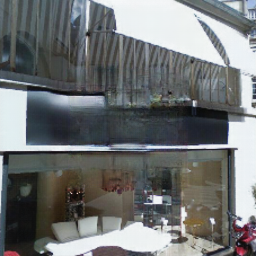} &
\includegraphics[width=.24\linewidth]{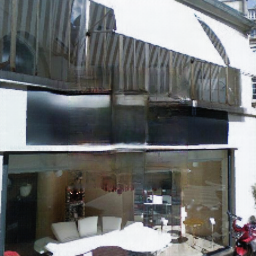} &
\includegraphics[width=.24\linewidth]{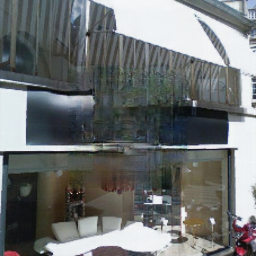} \\
(a) $\lambda_{g}=1$ & (b) $\lambda_{g}=0.1$ & (c) $\lambda_{g}=0.01$ & (d) $\lambda_{g}=0.001$\\
\end{tabular}
\vspace{-1em}
\caption{The effect of the tradeoff parameter $\lambda_{g}$ of guidance loss. }
\label{fig:lambdaG}
\vspace{-1em}
\end{figure}

The main differences between our Shift-Net and the other methods are the introduction of guidance loss and shift-connection layer.
Thus, experiments are first conducted to analyze the effect of guidance loss and shift operation.
Then we respectively zero out the corresponding weight of $(L-l+1)$-th layer to verify the effectiveness of the shift feature $\Phi_{L-l}^{\text{\emph{shift}}}$ in generating fine-detailed results.
Moreover, the benefit of shift-connection does not owe to the increase of feature map size.
To illustrate this, we also compare Shift-Net with a baseline model by substituting the nearest neighbor searching with random shift-connection.

\vspace{-1em}
\subsection{Effect of guidance loss}\label{section5.1}

Two groups of experiments are conducted to evaluate the effect of guidance loss.
In the first group of experiments, we add and remove the guidance loss ${\cal L}_g$ for U-Net and our Shift-Net to train the inpainting models.
Fig.~\ref{fig:guidanceEffectiveness} shows the inpainting results by these four methods.
It can be observed that, for both U-Net and Shift-Net the guidance loss is helpful in suppressing artifacts and preserving salient structure.

In the second group of experiments, we evaluate the effect of the tradeoff parameter $\lambda_g$ for guidance loss.
For our Shift-Net, the guidance loss is introduced for both recovering the semantic structure of the missing region and guiding the shift of the encoder feature.
To this end, proper tradeoff parameter $\lambda_g$ should be chosen, where too large or too small $\lambda_g$ values may be harmful to the inpainting results.
Fig.~\ref{fig:lambdaG} shows the results by setting different $\lambda_g$ values.
When $\lambda_{g}$ is small (e.g., $= 0.001$), the decoder feature may not serve as a suitable guidance to guarantee the correct shift of the encoder feature.
From Fig.~\ref{fig:lambdaG}(d), some artifacts can still be observed.
When $\lambda_{g}$ becomes too large (e.g., $\geq 0.1$), the constraint will be too excessive, and artifacts may also be introduced in the result (see Fig.~\ref{fig:lambdaG}(a)(b)).
Thus, we empirically set $\lambda_{g}=0.01$ in all our experiments.

\vspace{-1em}
\subsection{Effect of shift operation at different layers}\label{section5.2}

\begin{figure}[!t]
\setlength\tabcolsep{1.5pt}
\centering
\small
\begin{tabular}{cccc}
\includegraphics[width=.24\linewidth]{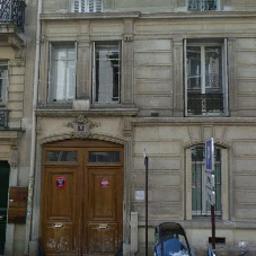} &
\includegraphics[width=.24\linewidth]{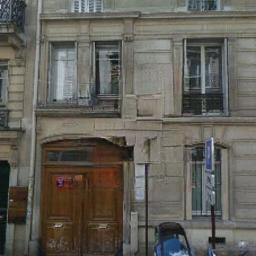} &
\includegraphics[width=.24\linewidth]{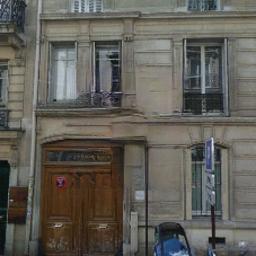} &
\includegraphics[width=.24\linewidth]{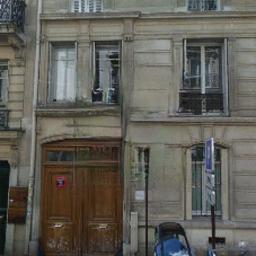} \\
(a) ground-truth & (b) $L-4$ & (c) $L-3$ & (d) $L-2$\\
\end{tabular}
\vspace{-1em}
\caption{The effect of performing shift operation on different layers $L-l$. }
\label{fig:Shift_in_layers}
\vspace{-1.5em}
\end{figure}

The superiority of Shift-Net against context encoder~\cite{pathak2016context} has demonstrated the effectiveness of shift operation.
By comparing the results by U-Net (w/${\cal L}_g$) and Shift-Net (w/${\cal L}_g$) in Fig.~\ref{fig:guidanceEffectiveness}(b)(d), one can see that shift operation does benefit the preserving of semantics and the recovery of detailed textures.
Note that the shift operation can be deployed to different layer, e.g., $(L-l)$-th, of the decoder.
When $l$ is smaller, the feature map size goes larger, and more computation time is required to perform the shift operation.
When $l$ is larger, the feature map size becomes smaller, but more detailed information may lost in the corresponding encoder layer, which may be harmful to recover image details and semantics.
Thus, proper $l$ should be chosen for better tradeoff between computation time and inpainting performance.
Fig.~\ref{fig:Shift_in_layers} shows the results of Shift-Net by adding the shift-connection layer to each of the $(L-4)$-th, $(L-3)$-th, and $(L-2)$-th layers, respectively.
When the shift-connection layer is added to the $(L-2)$-th layer, Shift-Net generally works well in producing visually pleasing results, but it takes more time (i.e., $\sim400$ ms per image) to process an image (See Fig.~\ref{fig:Shift_in_layers}(d)).
When the shift-connection layer is added to the $(L-4)$-th layer, Shift-Net becomes very efficient (i.e., $\sim40$ ms per image) but tends to generate the result with less textures and coarse details (See Fig.~\ref{fig:Shift_in_layers}(b)).
By performing the shift operation in $(L-3)$-th layer, better tradeoff between efficiency (i.e., $\sim80$ ms per image) and performance can be obtained by Shift-Net (See Fig.~\ref{fig:Shift_in_layers}(c)).

%%%%%%%%%%%%%%%%%%%%%%%%%%%%%%%%%%%%%%%%%%%%%%%%%%%%%%%%%%%%%%
%%%%%%%%%%%%%%% Figure Shift-connection after different layers
%%%%%%%%%%%%%%%%%%%%%%%%%%%%%%%%%%%%%%%%%%%%%%%%%%%%%%%%%%%%%%

\vspace{-1em}
\subsection{Effect of the shifted feature}

\begin{figure}[!t]
\setlength\tabcolsep{1.5pt}
\centering
\small
\begin{tabular}{ccccc}
\includegraphics[width=.18\linewidth]{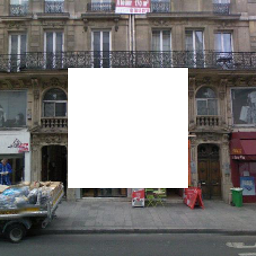} &
\includegraphics[width=.18\linewidth]{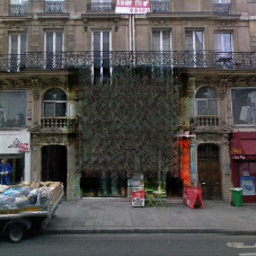} &
\includegraphics[width=.18\linewidth]{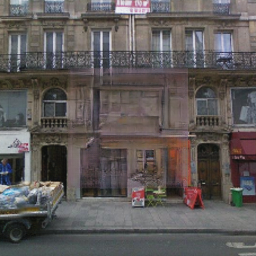} &
\includegraphics[width=.18\linewidth]{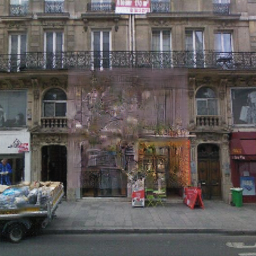} &
\includegraphics[width=.18\linewidth]{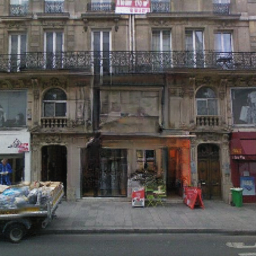} \\
(a)  & (b) & (c) & (d)& (e) \\
\end{tabular}
\vspace{-1em}
\caption{Given (a) the input, (b), (c) and (d) are respectively the results when the 1st, 2nd, 3rd parts of weights in $(L-l+1)$-th layer are zeroed. (e) is the result of Ours.}
\label{fig:Effect_of_shifted_feature}
\end{figure}

As we stacks the convolutional features $\Phi_{L-l}(I)$, $\Phi_{l}(I)$ and $\Phi_{L-l}^{\text{\emph{shift}}}$ as inputs
of $(L-l+1)$-th layer of U-Net, we can respectively zero out the weight of the corresponding slice in $(L-l+1)$-th layer.
%
% In this way, we can figure out the functionality of each part of features in our Shift-Net.
%
Fig.~\ref{fig:Effect_of_shifted_feature} demonstrates the results of Shift-Net by only zeroing out the weight of each slice.
When we abandon the decoder feature $\Phi_{L-l}(I)$, the central part fails to restore any structures (See Fig.~\ref{fig:Effect_of_shifted_feature}(b)), which indicates main structure and content are constructed by the subnet between
$l\sim L-l$ layers.
However, if we ignore the feature $\Phi_{l}(I)$, we get general structure (See (Fig.~\ref{fig:Effect_of_shifted_feature}(c)) but quality inferior to the final result Fig.~\ref{fig:Effect_of_shifted_feature}(e).
This exhibits the fact that encoder feature $\Phi_{l}(I)$ has no significant effect on recovering features, which
manifests the guidance loss is forceful to explicitly model the relationship between $\left( \Phi_{L-l}(I) \right)_{\mathbf{y}}$ and $\left( \Phi_{l}(I^{gt}) \right)_{\mathbf{y}}$ as illustrated in Sec.~\ref{section3.1}.
Finally, when we discard the shift feature $\Phi_{L-l}^{\text{\emph{shift}}}$, the result is totally a mixture of structures
(See Fig.~\ref{fig:Effect_of_shifted_feature}(d)).
Therefore, we can conclude that $\Phi_{L-l}^{\text{\emph{shift}}}$ acts as a refinement and enhancement role in recovering clear and fine details
in our Shift-Net.

%%%%%%%%%%%%%%%%%%%%%%%%%%%%%%%%%%%%%%%%%
%%% Figure Effect of shifted feature
%%%%%%%%%%%%%%%%%%%%%%%%%%%%%%%%%%%%%%%%%

\begin{figure}[!t]
  \centering
\setlength\tabcolsep{1.5pt}
\begin{tabular}{ccccc}
  \includegraphics[width=.18\linewidth]{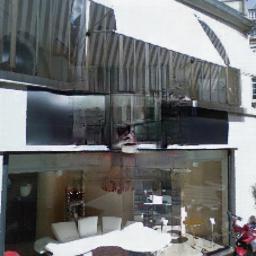}&
  \includegraphics[width=.18\linewidth]{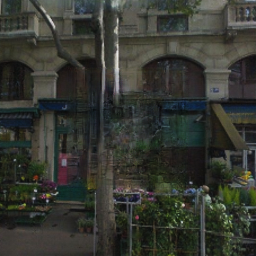}&
  \includegraphics[width=.18\linewidth]{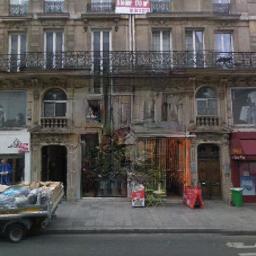}&
  \includegraphics[width=.18\linewidth]{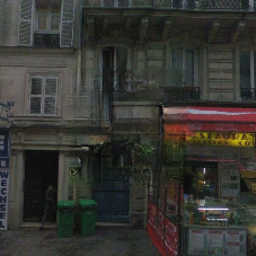}&
  \includegraphics[width=.18\linewidth]{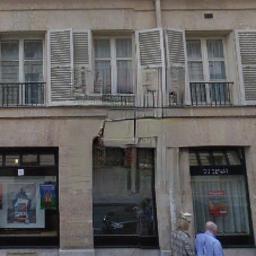}\\

  \includegraphics[width=.18\linewidth]{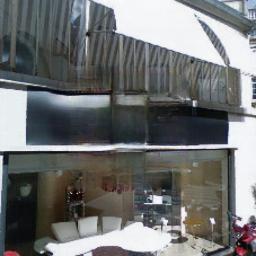}&
  \includegraphics[width=.18\linewidth]{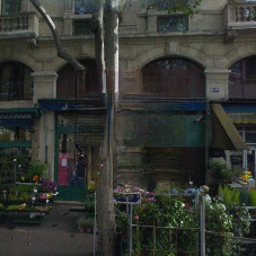}&
  \includegraphics[width=.18\linewidth]{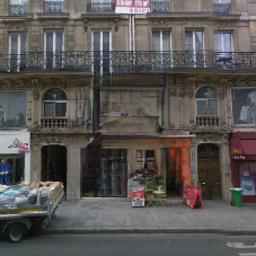}&
  \includegraphics[width=.18\linewidth]{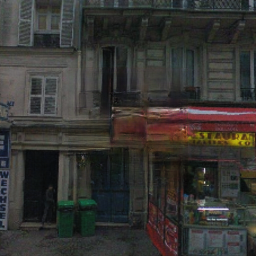}&
  \includegraphics[width=.18\linewidth]{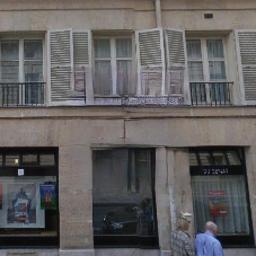}\\

\end{tabular}
\vspace{-1em}
\caption{From top to bottom are: Shift-Net with random shift-connection and nearest neighbor searching.}
\label{fig:Comparison_randomReplace}
\vspace{-1em}
\end{figure}

\vspace{-1em}
\subsection{Comparison with random shift-connection}\label{setcion5.4}
Finally, we implement a baseline Shift-Net model by substituting the nearest neighbor searching with random shift-connection.
Fig.~\ref{fig:Comparison_randomReplace} shows five examples of inpainting results by Shift-Net and baseline model.
Compared to the nearest neighbor searching, the results by random shift-connection exhibit more artifacts, distortions, and structure disconnections.
When training with random neighbor searching, random shifted feature continuously acts as dummy and confusing input.
The network gradually learns to ignore $\Phi_{L-l}^{\text{\emph{shift}}}$ in order to minimizing the total loss function.
Thus, the favorable performance of Shift-Net should owe to the correct shift-operation.

\vspace{-1em}
\section{Conclusion}\label{section6}

This paper has proposed a novel architecture, i.e., Shift-Net, for image completion that exhibits fast speed with promising fine details via deep feature rearrangement.
The guidance loss is introduced to enhance the explicit relation between the encoded feature in known region and decoded feature in missing region.
By exploiting such relation, the shift operation can be efficiently performed and is effective in improving inpainting performance.
Experiments show that our Shift-Net performs favorably in comparison to the state-of-the-art methods, and is effective in generating sharp, fine-detailed and photo-realistic images.
In future, more studies will be given to improve the speed of nearest searching in the shift operation, introduce multiple shift-connection layers, and extend the shift-connection to other low level vision tasks.

\clearpage
\appendix

%------------------- section B -----------------------------
\section{Definition of masked region in feature maps}\label{sectionA}
\begin{figure}[!htb]
\setlength\tabcolsep{1.5pt}
\centering
\small
\begin{tabular}{cccc}
\includegraphics[width=.22\linewidth]{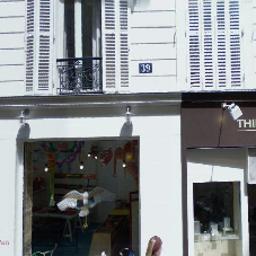} &
\includegraphics[width=.22\linewidth]{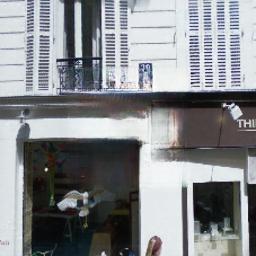} &
\includegraphics[width=.22\linewidth]{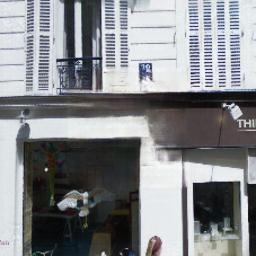} &
\includegraphics[width=.22\linewidth]{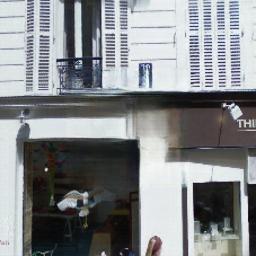} \\
(a) Ground-truth & (b) $T = 4/16$ & (c) $T = 5/16$ & (d) $T = 6/16$\\
\end{tabular}
\caption{The effect of different thresholds in shift-connection. }
\vspace{-2em}
\label{fig:Threshold_in_shift}
\vspace{-0.5em}
\end{figure}

As shift-connection works based on the boundary of masked region and unmasked region in feature maps.
Thus, we need to give a definition of masked region in feature maps.
Denote by $\Omega^0$ the missing part of the input image, and we should determine $\Omega^l$ for the $l$-th convolutional layer.
In our implementation, we introduce a mask image $M$ with $(M)_{\mathbf{y}} = 1$ ($\mathbf{y} \in \Omega$) and 0 otherwise.
Then, we define a CNN $\Psi(M)$ that has the same architecture with the encoder but with the network width of 1.
All the elements of the filters are $1/16$, and we remove all the nonlinearity.
Taking $M$ as input, we obtain the feature of the $l$-th layer as $\Psi_l(M)$.
Then, $\Omega^l$ is defined as $\Omega^l = \{ \mathbf{y} | (\Psi_l(M))_{\mathbf{y}} \geq T\}$, where $T$ is the threshold with $0 \leq T \leq 1$.
Fig.~\ref{fig:Threshold_in_shift} shows the results of Shift-Net by setting $T = 4/16, 5/16, 6/16$, respectively.
It can be seen that Shift-Net is robust to $T$, which may be attributed to that we take the shift and encoder, decoder features as the inputs to the $L-l+1$ layer.
We empirically set $T=5/16$ in our experiments.

%%%%%%%%%%%%%%%%%%%%%%%%%%%%%%%%%%%%%%%%%
%%% Figure Effect of different thresholds
%%%%%%%%%%%%%%%%%%%%%%%%%%%%%%%%%%%%%%%%%

%------------------------------------------------------------------------
\section{Details on Shift-Net}\label{setionB}

\subsection{Architecture of generative model G.}

%\noindent\textbf{}
For the generative model of our Shift-Net, we adopt the architecture of U-Net proposed in~\cite{isola2016image, radford2015unsupervised}.
Each convolution or deconvolution layer is followed by instance normalization~\cite{ulyanov2016instance}.
The encoder part of $G$ is stacked with Convolution-InstanceNorm-LeakyReLU layers, while the decoder part of $G$ consists of seven Deconvolution-InstanceNorm-ReLU layers.
Following the code of pix2pix, we zero out the biases of all convolution and deconvolution layers in the generative model in training.
In this way, we can promise the correctness of \textbf{Line 208}.
$L$ denotes the total number of convolution/deconvolution layers in our model.
We add guidance loss and shift operation in $(L-3)$-th layer, which results in the concatenated features of $\Phi_{L-3}(I)$, $\Phi_{3}(I)$ and $\Phi_{L-3}^{\text{\emph{shift}}}(I)$ as inputs of the adjacent deconvolution.
Details about the architecture of our generative model $G$ is shown in Table~\ref{table:netG}.
It is noted that we do not apply InstanceNorm on the bottleneck layer.
The activation map of the bottleneck layer is $1 \times 1$, which means we only get one activation per convolutional filter.
As we train our network with batchsize 1, activations will be zeroed out once InstanceNorm is applied on the bottleneck layer. Please to pix2pix\footnote{https://github.com/phillipi/pix2pix/commit/b479b6b} for more explanation.

\subsection{Architecture of discriminative network D.}\label{sectionC}

$D$ shares the similar design pattern with the encoder part of $G$, however, is only 5-convolution-layer network.
We exclusively use convolution layers with filters of size $4 \times 4$ pixels with varying stride lengths to reduce
the spatial dimension of the input down to a size of $30 \times 30$ where we append sigmoid activation at the final output.
InstanceNorm is not applied to the first convolutional layer, and we use leaky ReLU with slope of 0.2 for activations except
for the sigmoid in the last layer.
See Table~\ref{table:netD} for more details.

\section{More comparisons and object removals}\label{setionD}

\subsection{Comparisons on Paris StreetView and Places datasets}

\begin{figure*}[t]
  \center
\setlength\tabcolsep{0.5pt}
\begin{tabular}{cccccc}

  \includegraphics[width=.160\textwidth]{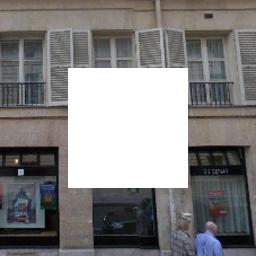}&
  \includegraphics[width=.160\textwidth]{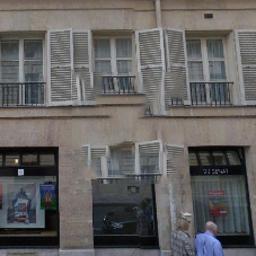}&
  \includegraphics[width=.160\textwidth]{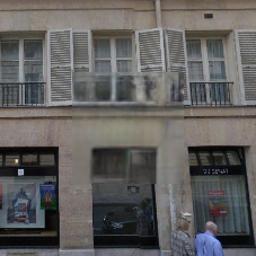}&
  \includegraphics[width=.160\textwidth]{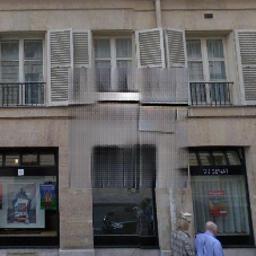}&
  \includegraphics[width=.160\textwidth]{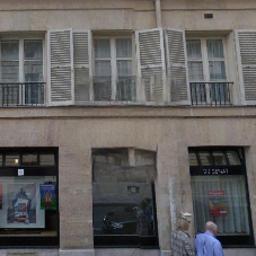}&
  \includegraphics[width=.160\textwidth]{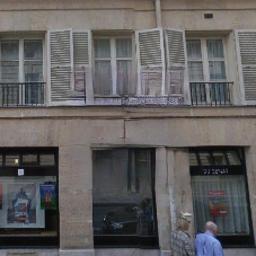}\\

  \includegraphics[width=.160\textwidth]{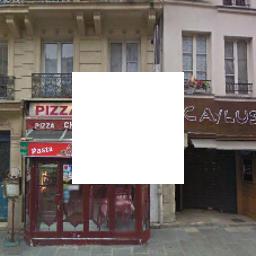}&
  \includegraphics[width=.160\textwidth]{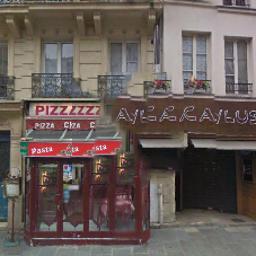}&
  \includegraphics[width=.160\textwidth]{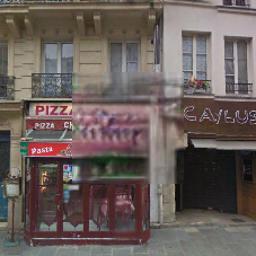}&
  \includegraphics[width=.160\textwidth]{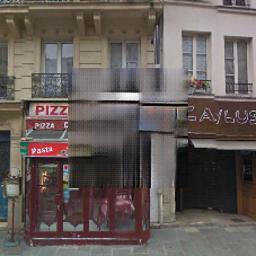}&
  \includegraphics[width=.160\textwidth]{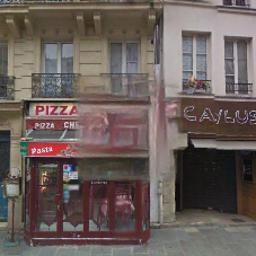}&
  \includegraphics[width=.160\textwidth]{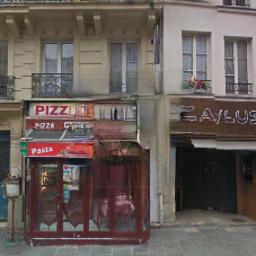}\\

  \includegraphics[width=.160\textwidth]{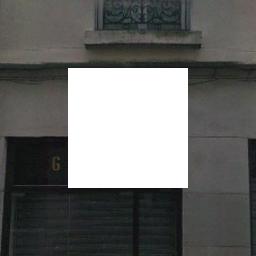}&
  \includegraphics[width=.160\textwidth]{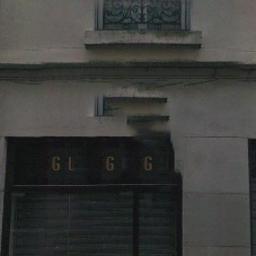}&
  \includegraphics[width=.160\textwidth]{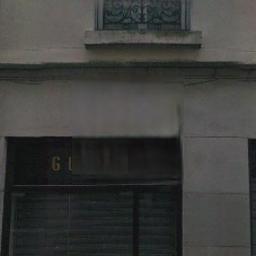}&
  \includegraphics[width=.160\textwidth]{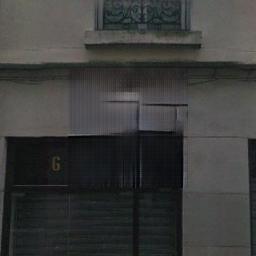}&
  \includegraphics[width=.160\textwidth]{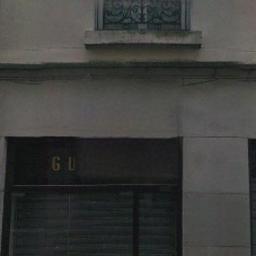}&
  \includegraphics[width=.160\textwidth]{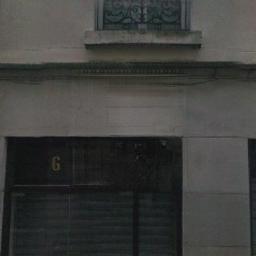}\\

  \includegraphics[width=.160\textwidth]{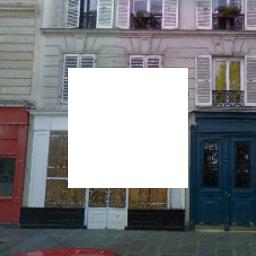}&
  \includegraphics[width=.160\textwidth]{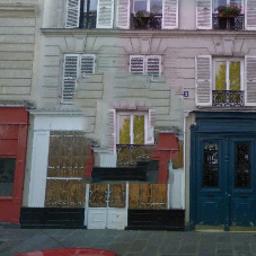}&
  \includegraphics[width=.160\textwidth]{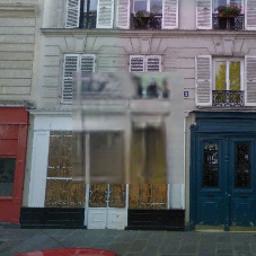}&
  \includegraphics[width=.160\textwidth]{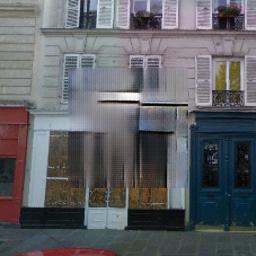}&
  \includegraphics[width=.160\textwidth]{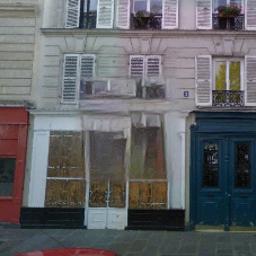}&
  \includegraphics[width=.160\textwidth]{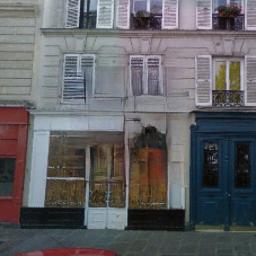}\\

    \includegraphics[width=.160\textwidth]{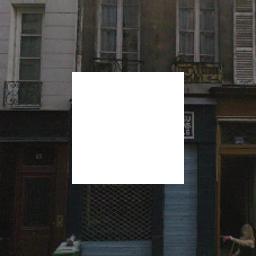}&
  \includegraphics[width=.160\textwidth]{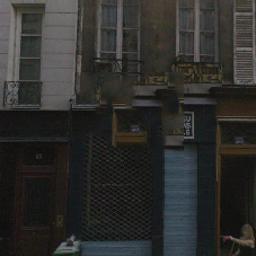}&
  \includegraphics[width=.160\textwidth]{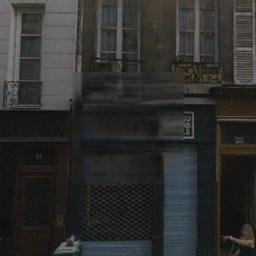}&
  \includegraphics[width=.160\textwidth]{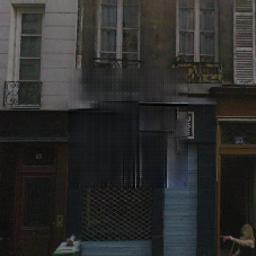}&
  \includegraphics[width=.160\textwidth]{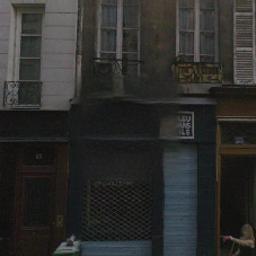}&
  \includegraphics[width=.160\textwidth]{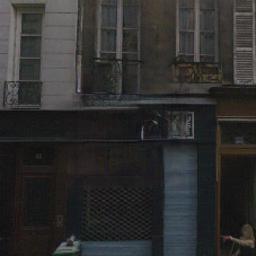}\\

  \includegraphics[width=.160\textwidth]{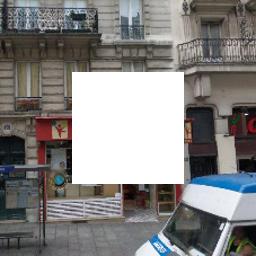}&
  \includegraphics[width=.160\textwidth]{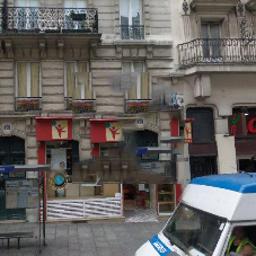}&
  \includegraphics[width=.160\textwidth]{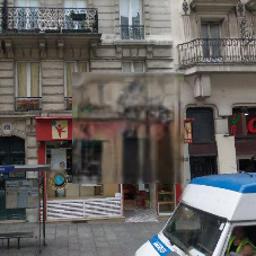}&
  \includegraphics[width=.160\textwidth]{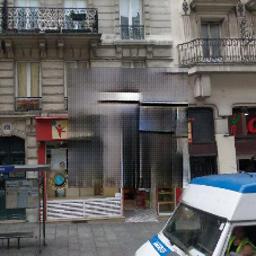}&
  \includegraphics[width=.160\textwidth]{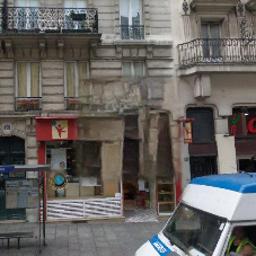}&
  \includegraphics[width=.160\textwidth]{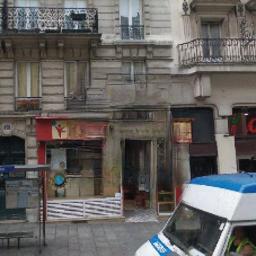}\\

  \includegraphics[width=.160\textwidth]{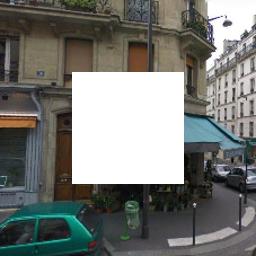}&
  \includegraphics[width=.160\textwidth]{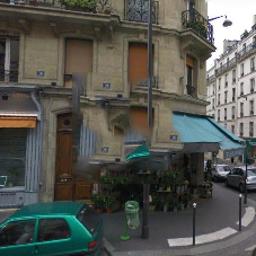}&
  \includegraphics[width=.160\textwidth]{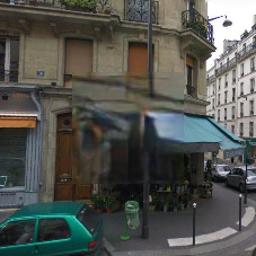}&
  \includegraphics[width=.160\textwidth]{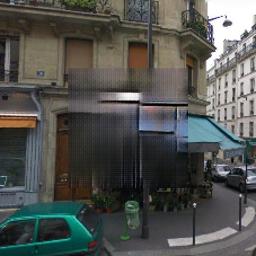}&
  \includegraphics[width=.160\textwidth]{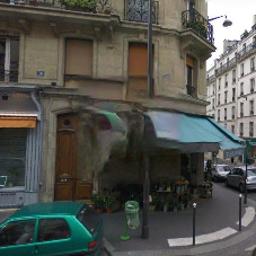}&
  \includegraphics[width=.160\textwidth]{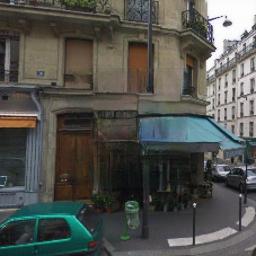}\\

  \includegraphics[width=.160\textwidth]{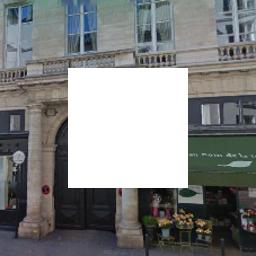}&
  \includegraphics[width=.160\textwidth]{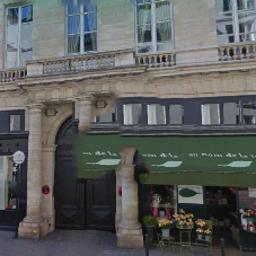}&
  \includegraphics[width=.160\textwidth]{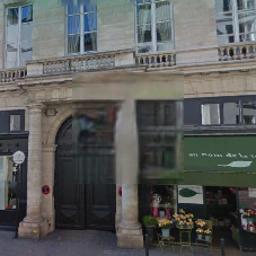}&
  \includegraphics[width=.160\textwidth]{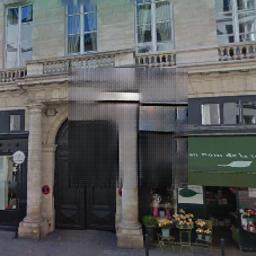}&
  \includegraphics[width=.160\textwidth]{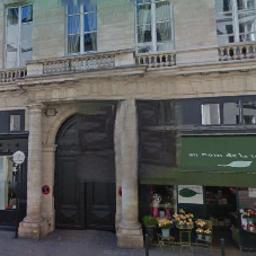}&
  \includegraphics[width=.160\textwidth]{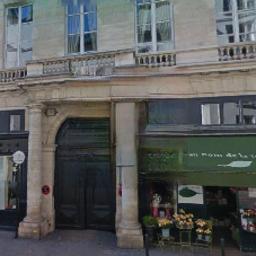}\\

(a)  & (b) & (c)  & (d) & (e) & (f)\\
\end{tabular}
\vspace{-.5em}
\caption{Qualitative comparisons on Paris StreetView. From the left to the right are:
(a) input, (b) Content-Aware Fill~\cite{Content-Aware-Fill}, (c) context encoder~\cite{pathak2016context}, (d) pix2pix\cite{isola2016image}, (e) MNPS~\cite{yang2017high} and (f) Ours. All images are scaled to $256\times 256$.}

\label{fig:comparison_on_paris_1}
\vspace{-.5em}
\end{figure*}

\begin{figure*}[!htbp]
  \center
\setlength\tabcolsep{0.5pt}
\begin{tabular}{cccccc}

  \includegraphics[width=.160\textwidth]{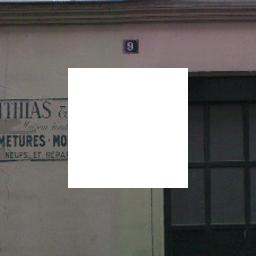}&
  \includegraphics[width=.160\textwidth]{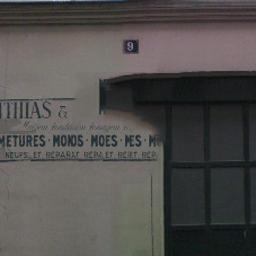}&
  \includegraphics[width=.160\textwidth]{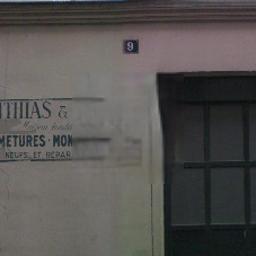}&
  \includegraphics[width=.160\textwidth]{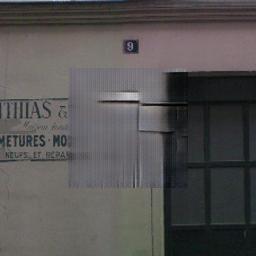}&
  \includegraphics[width=.160\textwidth]{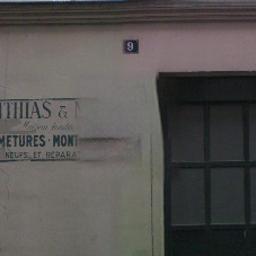}&
  \includegraphics[width=.160\textwidth]{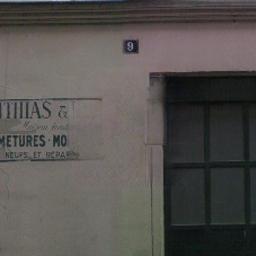}\\

  \includegraphics[width=.160\textwidth]{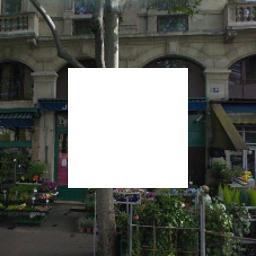}&
  \includegraphics[width=.160\textwidth]{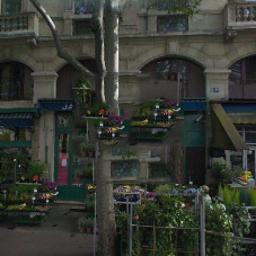}&
  \includegraphics[width=.160\textwidth]{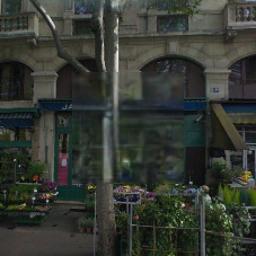}&
  \includegraphics[width=.160\textwidth]{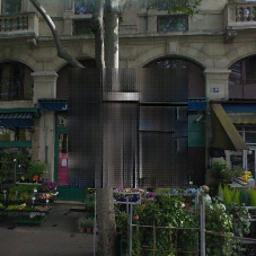}&
  \includegraphics[width=.160\textwidth]{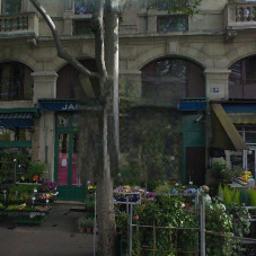}&
  \includegraphics[width=.160\textwidth]{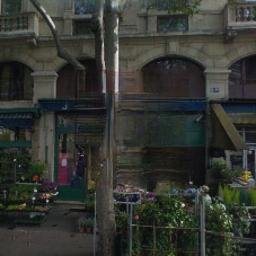}\\

    \includegraphics[width=.160\textwidth]{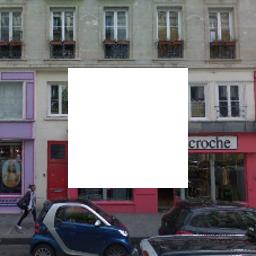}&
  \includegraphics[width=.160\textwidth]{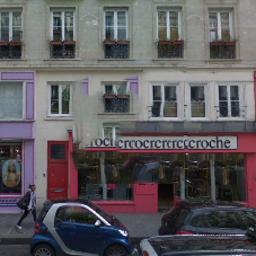}&
  \includegraphics[width=.160\textwidth]{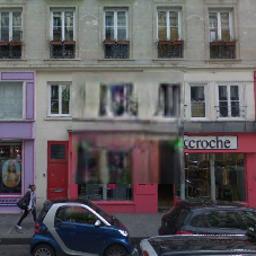}&
  \includegraphics[width=.160\textwidth]{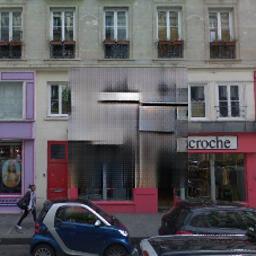}&
  \includegraphics[width=.160\textwidth]{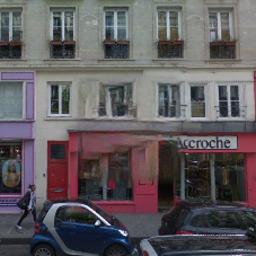}&
  \includegraphics[width=.160\textwidth]{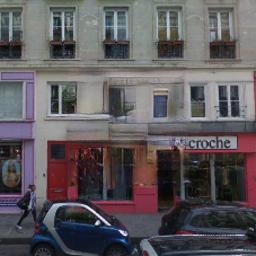}\\

  \includegraphics[width=.160\textwidth]{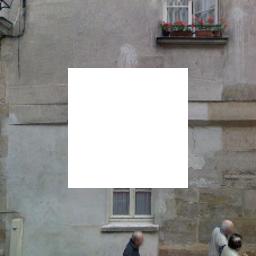}&
  \includegraphics[width=.160\textwidth]{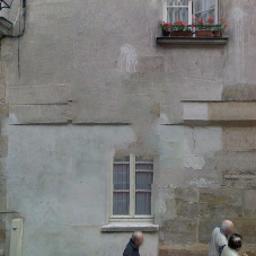}&
  \includegraphics[width=.160\textwidth]{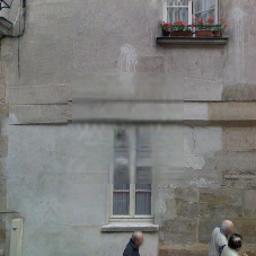}&
  \includegraphics[width=.160\textwidth]{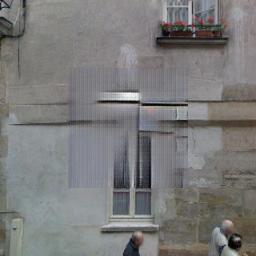}&
  \includegraphics[width=.160\textwidth]{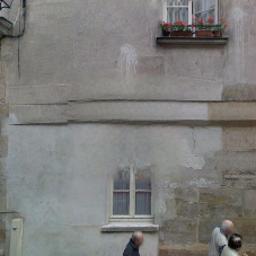}&
  \includegraphics[width=.160\textwidth]{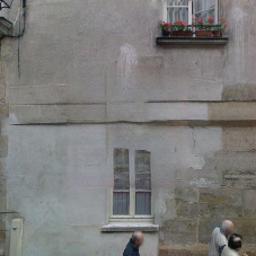}\\

  \includegraphics[width=.160\textwidth]{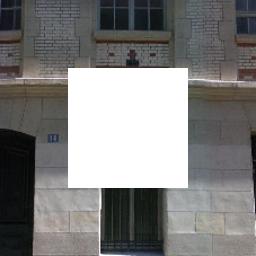}&
  \includegraphics[width=.160\textwidth]{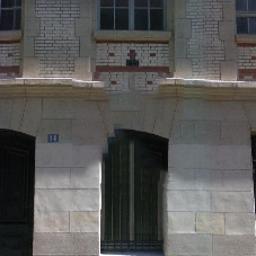}&
  \includegraphics[width=.160\textwidth]{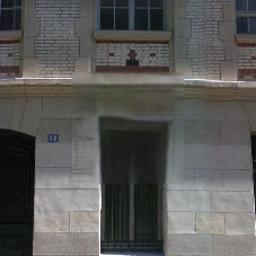}&
  \includegraphics[width=.160\textwidth]{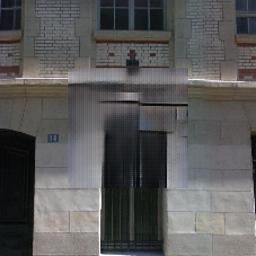}&
  \includegraphics[width=.160\textwidth]{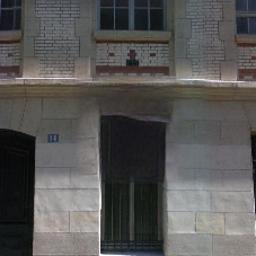}&
  \includegraphics[width=.160\textwidth]{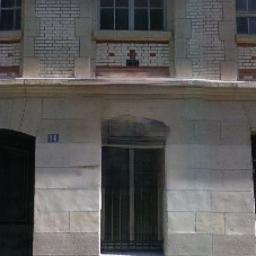}\\

  \includegraphics[width=.160\textwidth]{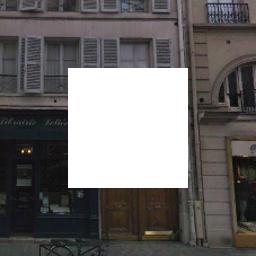}&
  \includegraphics[width=.160\textwidth]{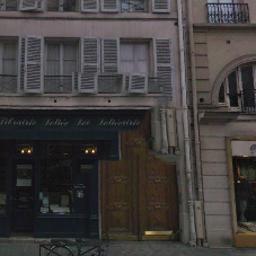}&
  \includegraphics[width=.160\textwidth]{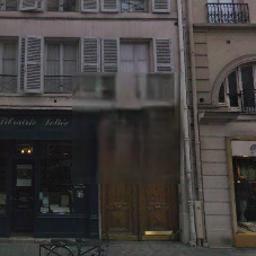}&
  \includegraphics[width=.160\textwidth]{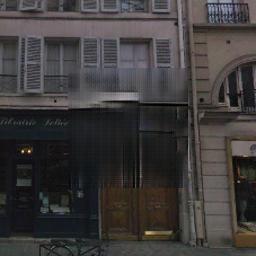}&
  \includegraphics[width=.160\textwidth]{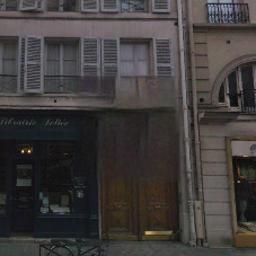}&
  \includegraphics[width=.160\textwidth]{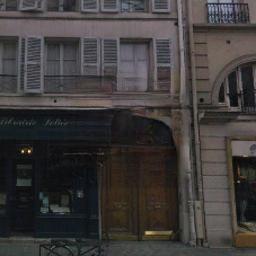}\\

  \includegraphics[width=.160\textwidth]{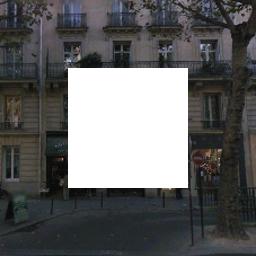}&
  \includegraphics[width=.160\textwidth]{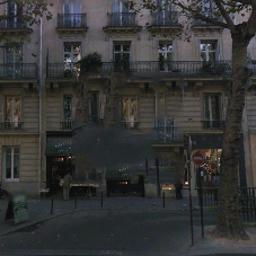}&
  \includegraphics[width=.160\textwidth]{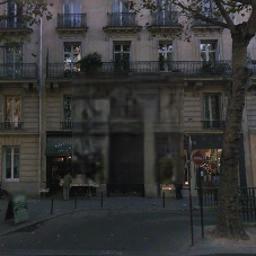}&
  \includegraphics[width=.160\textwidth]{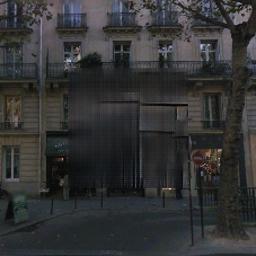}&
  \includegraphics[width=.160\textwidth]{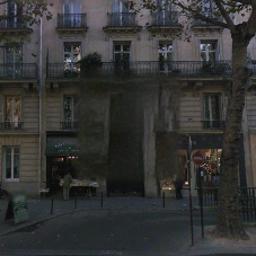}&
  \includegraphics[width=.160\textwidth]{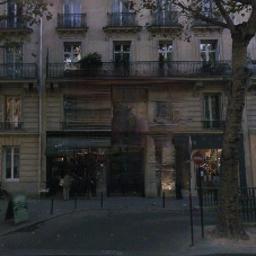}\\

  \includegraphics[width=.160\textwidth]{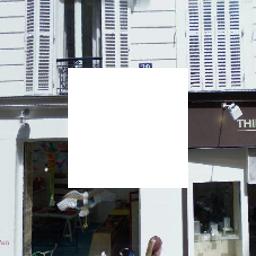}&
  \includegraphics[width=.160\textwidth]{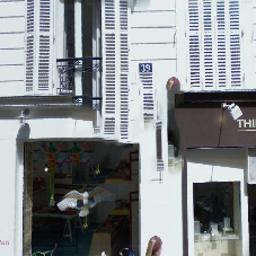}&
  \includegraphics[width=.160\textwidth]{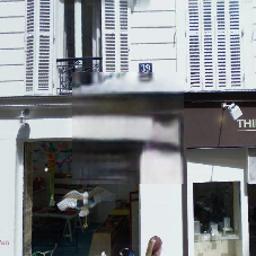}&
  \includegraphics[width=.160\textwidth]{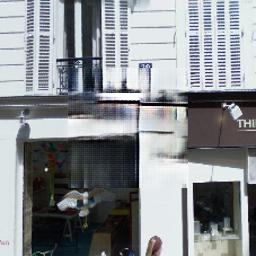}&
  \includegraphics[width=.160\textwidth]{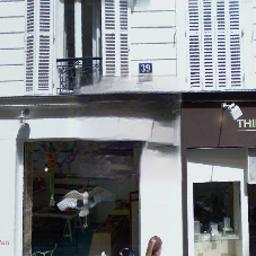}&
  \includegraphics[width=.160\textwidth]{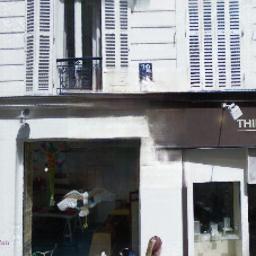}\\

(a)  & (b) & (c)  & (d) & (e) & (f)\\
\end{tabular}
\vspace{-.5em}
\caption{Qualitative comparisons on Paris StreetView. From the left to the right are:
(a) input, (b) Content-Aware Fill~\cite{Content-Aware-Fill}, (c) context encoder~\cite{pathak2016context}, (d) pix2pix\cite{isola2016image}, (e) MNPS~\cite{yang2017high} and (f) Ours. All images are scaled to $256\times 256$.}

\label{fig:comparison_on_paris_2}
\vspace{-.5em}
\end{figure*}

\begin{figure*}[t]
  \center
\setlength\tabcolsep{0.5pt}
\begin{tabular}{cccccc}
  \includegraphics[width=.160\textwidth]{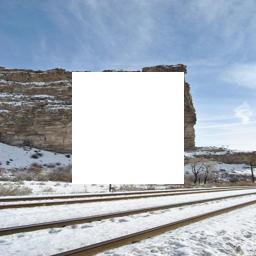}&
  \includegraphics[width=.160\textwidth]{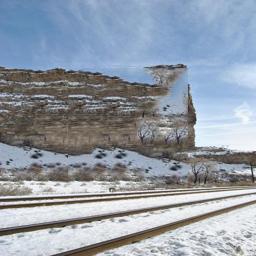}&
  \includegraphics[width=.160\textwidth]{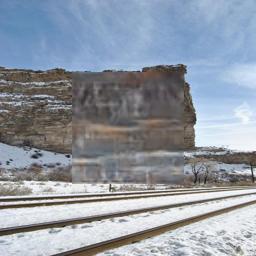}&
  \includegraphics[width=.160\textwidth]{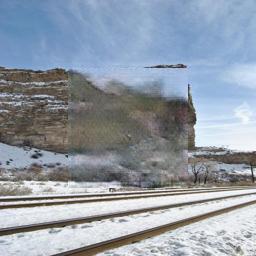}&
  \includegraphics[width=.160\textwidth]{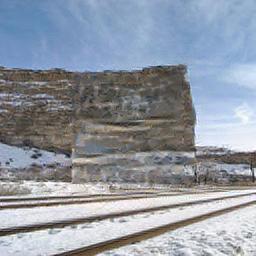}&
  \includegraphics[width=.160\textwidth]{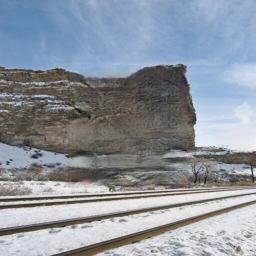}\\

  \includegraphics[width=.160\textwidth]{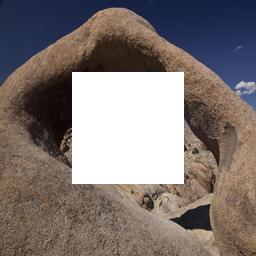}&
  \includegraphics[width=.160\textwidth]{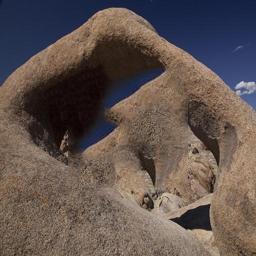}&
  \includegraphics[width=.160\textwidth]{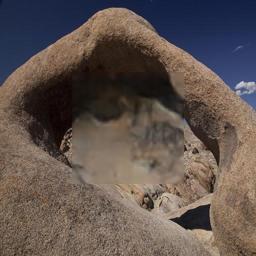}&
  \includegraphics[width=.160\textwidth]{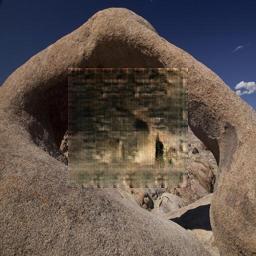}&
  \includegraphics[width=.160\textwidth]{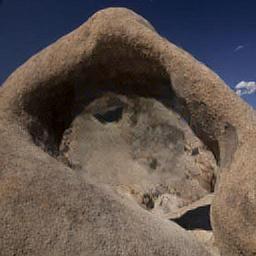}&
  \includegraphics[width=.160\textwidth]{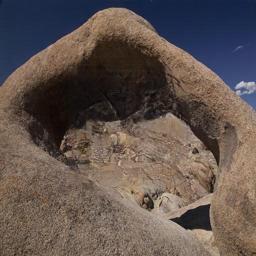}\\

  \includegraphics[width=.160\textwidth]{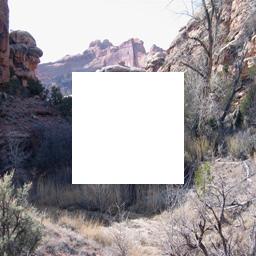}&
  \includegraphics[width=.160\textwidth]{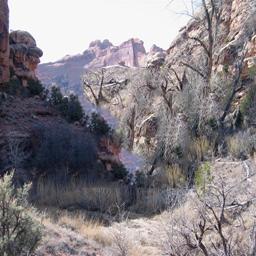}&
  \includegraphics[width=.160\textwidth]{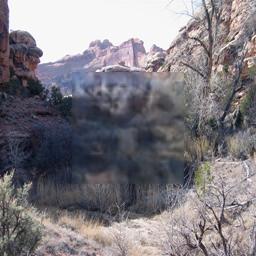}&
  \includegraphics[width=.160\textwidth]{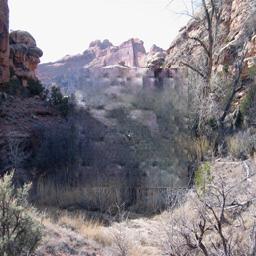}&
  \includegraphics[width=.160\textwidth]{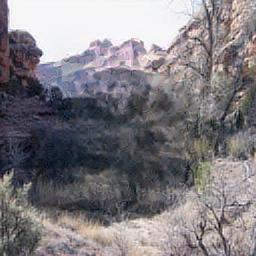}&
  \includegraphics[width=.160\textwidth]{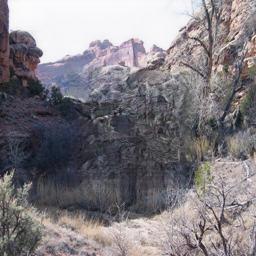}\\

  \includegraphics[width=.160\textwidth]{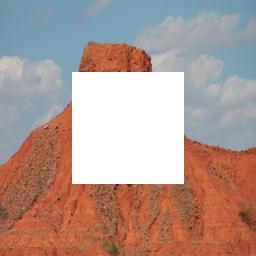}&
  \includegraphics[width=.160\textwidth]{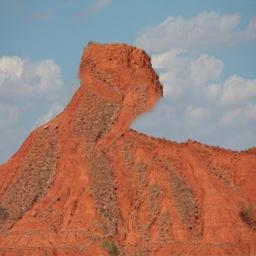}&
  \includegraphics[width=.160\textwidth]{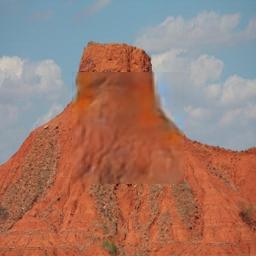}&
  \includegraphics[width=.160\textwidth]{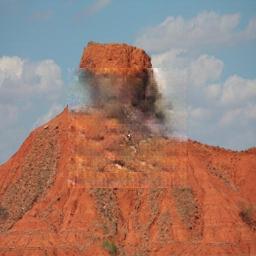}&
  \includegraphics[width=.160\textwidth]{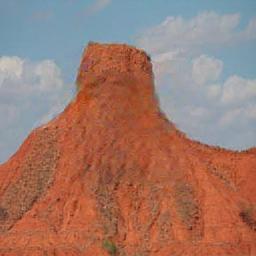}&
  \includegraphics[width=.160\textwidth]{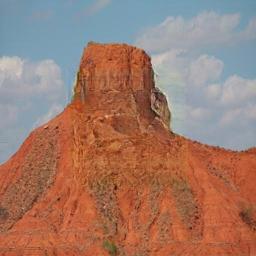}\\

  \includegraphics[width=.160\textwidth]{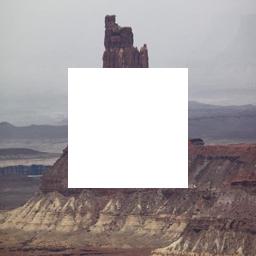}&
  \includegraphics[width=.160\textwidth]{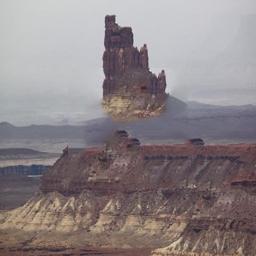}&
  \includegraphics[width=.160\textwidth]{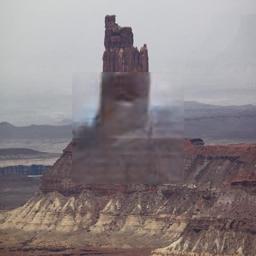}&
  \includegraphics[width=.160\textwidth]{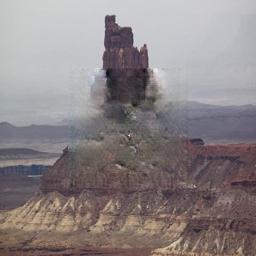}&
  \includegraphics[width=.160\textwidth]{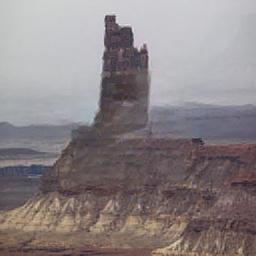}&
  \includegraphics[width=.160\textwidth]{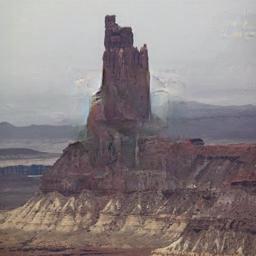}\\

  \includegraphics[width=.160\textwidth]{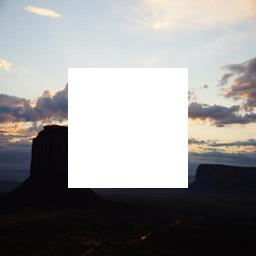}&
  \includegraphics[width=.160\textwidth]{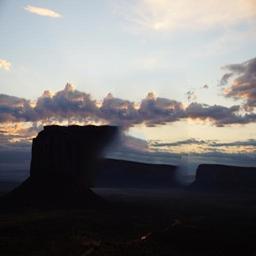}&
  \includegraphics[width=.160\textwidth]{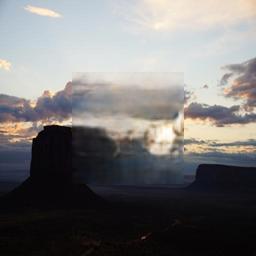}&
  \includegraphics[width=.160\textwidth]{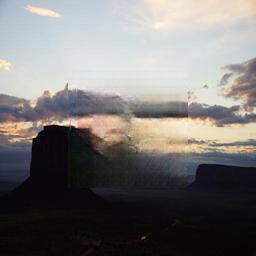}&
  \includegraphics[width=.160\textwidth]{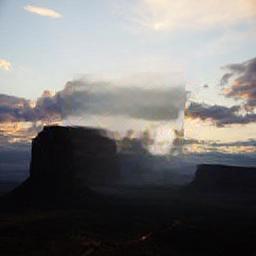}&
  \includegraphics[width=.160\textwidth]{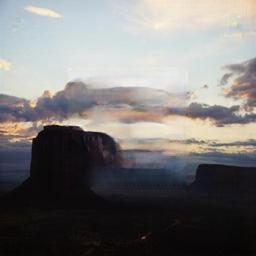}\\

  \includegraphics[width=.160\textwidth]{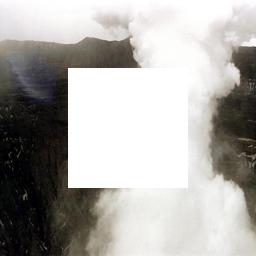}&
  \includegraphics[width=.160\textwidth]{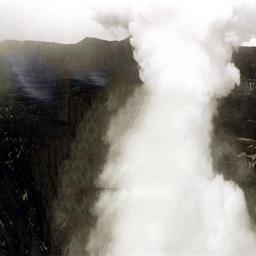}&
  \includegraphics[width=.160\textwidth]{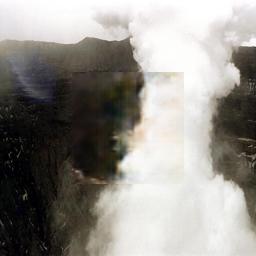}&
  \includegraphics[width=.160\textwidth]{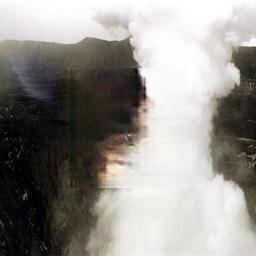}&
  \includegraphics[width=.160\textwidth]{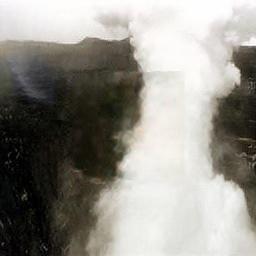}&
  \includegraphics[width=.160\textwidth]{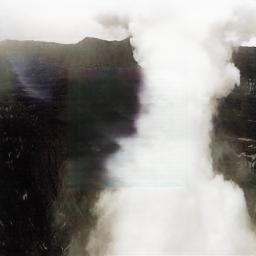}\\

  \includegraphics[width=.160\textwidth]{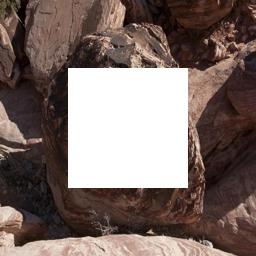}&
  \includegraphics[width=.160\textwidth]{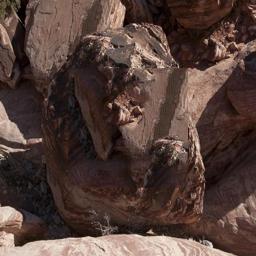}&
  \includegraphics[width=.160\textwidth]{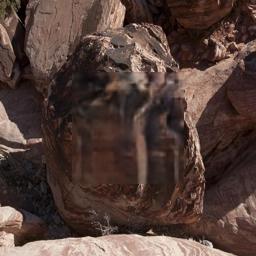}&
  \includegraphics[width=.160\textwidth]{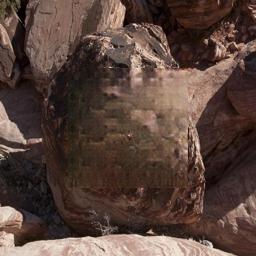}&
  \includegraphics[width=.160\textwidth]{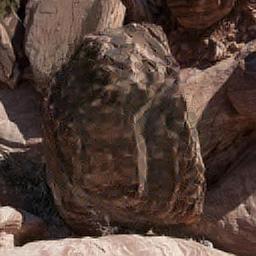}&
  \includegraphics[width=.160\textwidth]{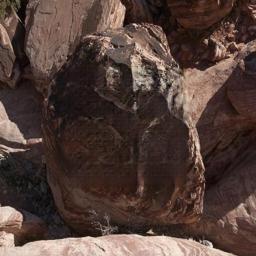}\\

(a)  & (b) & (c)  & (d) & (e) & (f)\\
\end{tabular}
\vspace{-.5em}
\caption{Qualitative comparisons on Places. From the left to the right are:
(a) input, (b) Content-Aware Fill~\cite{Content-Aware-Fill}, (c) context encoder~\cite{pathak2016context}, (d) pix2pix\cite{isola2016image}, (e) MNPS~\cite{yang2017high} and (f) Ours. All images are scaled to $256\times 256$.}

\label{fig:comparison_on_place_1}
\vspace{-.5em}
\end{figure*}

More comparisons with context encoder~\cite{pathak2016context}, Content-Aware-Fill~\cite{Content-Aware-Fill}, pix2pix\cite{isola2016image} and MNPS~\cite{yang2017high} on both Paris StreetView~\cite{doersch2012makes} and
Places\cite{zhou2017places} are also conducted.
Please refer to Fig.~\ref{fig:comparison_on_paris_1} and~\ref{fig:comparison_on_paris_2} for more results on Paris StreetView. For comparison on Places, please refer to Fig.~\ref{fig:comparison_on_place_1}.
Our Shift-Net outperforms state-of-the-art approaches in both structural consistency and detail richness.
Both global structure and fine details can be preserved in our model, however, other methods either perform poorly in generating clear, realistic details or lack global structure consistency.
%

%%%%%%%%%%%%%%%%%%%%%%%%%%%%%%%%%%%%%%%
%%%%   Table: netG
%%%%%%%%%%%%%%%%%%%%%%%%%%%%%%%%%%%%%%%

\begin{table}[!t]
\renewcommand{\arraystretch}{1.3}
\caption{The architecture of the $G$ network. ``IN'' represents InstanceNorm and ``LReLU''
donates leaky ReLU with the slope of 0.2.}
\vspace{-0.5em}
\centering
\begin{tabular}{l}
\hline
\ \ \ \ \ \ \ \ \ \ \ \ {\bf The architecture of generative model} $G$\\
\hline
{\bf Input}: Image ($256 \times 256 \times 3$)\\
\hline
[Layer \ \ 1]    \ \ \  Conv. (4, 4, \ 64), stride=2; \\  %e1
\hline
[Layer \ \ 2]   \ \ \ \emph{LReLU}; Conv. (4, 4, 128), stride=2; IN; \\ %e2
\hline
[Layer \ \ 3]    \ \ \ \emph{LReLU}; Conv. (4, 4, 256), stride=2; IN; \\ %e3
\hline
[Layer \ \ 4]    \ \ \ \emph{LReLU}; Conv. (4, 4, 512), stride=2; IN; \\ %e4
\hline
[Layer \ \ 5]    \ \ \ \emph{LReLU}; Conv. (4, 4, 512), stride=2; IN;\\ %e5
\hline
[Layer \ \ 6]    \ \ \ \emph{LReLU}; Conv. (4, 4, 512), stride=2; IN;\\ %e6
\hline
[Layer \ \ 7]    \ \ \ \emph{LReLU}; Conv. (4, 4, 512), stride=2; IN;\\ %e7
\hline
[Layer \ \ 8]    \ \ \ \emph{LReLU}; Conv. (4, 4, 512), stride=2; \\ %e8
\hline
[Layer \ \ 9]    \ \ \  \emph{ReLU}; DeConv. (4, 4, 512), stride=2; IN; \ \\ %d1_
\hline
\ \ \ \ \ \ \ \ \ \ \ \ \ \ \ \ \ \ \ \ \ Concatenate(Layer \ \ 9, Layer \ 7);\\
\hline
[Layer 10]    \ \ \  DeConv. (4, 4, 512), stride=2; IN; \ \\ %d2_
\hline
\ \ \ \ \ \ \ \ \ \ \ \ \ \ \ \ \ \ \ \ \ Concatenate(Layer 10, Layer \ 6); \emph{ReLU};\\
\hline
[Layer 11]    \ \ \  DeConv. (4, 4, 512), stride=2; IN; \ \\ %d3_
\hline
\ \ \ \ \ \ \ \ \ \ \ \ \ \ \ \ \ \ \ \ \ Concatenate(Layer 11, Layer \ 5); \emph{ReLU};\\
\hline
[Layer 12]    \ \ \  DeConv. (4, 4, 512), stride=2; IN; \ \\ %d4_
\hline
\ \ \ \ \ \ \ \ \ \ \ \ \ \ \ \ \ \ \ \ \ Concatenate(Layer 12, Layer \ 4); \emph{ReLU};\\
\hline
[Layer 13]    \ \ \  DeConv. (4, 4, 256), stride=2; IN; \ \\ %d5_
\hline
\ \ \ \ \ \ \ \ \ \ \ \ \ \ \ \ \ \ \ \ \ Concatenate(Layer 13, Layer \ 3); \emph{ReLU};\\
\hline
[Layer 14]    \ \ \  {\bf Guidance loss layer}; \ \\
\hline
[Layer 15]    \ \ \  {\bf Shift-connection layer}; \ \\
\hline
[Layer 16]    \ \ \  DeConv. (4, 4, 128), stride=2; IN; \ \\ %d6_
\hline
\ \ \ \ \ \ \ \ \ \ \ \ \ \ \ \ \ \ \ \ \ Concatenate(Layer 16, Layer \ 2); \emph{ReLU};\\
\hline
[Layer 17]    \ \ \  DeConv. (4, 4, \ \ 64), stride=2; IN; \\  %d7_
\hline
\ \ \ \ \ \ \ \ \ \ \ \ \ \ \ \ \ \ \ \ \ Concatenate(Layer 17, Layer \ 1); \emph{ReLU};\\
\hline
[Layer 18]    \ \ \  \emph{ReLU}; DeConv. (4, 4, 3), stride=2; \emph{Tanh}; \ \ \  \\
\hline
{\bf Output}: Final result ($256 \times 256 \times 3$)\\
\hline
\end{tabular}
\label{table:netG}
\end{table}

%%%%%%%%%%%%%%%%%%%%%%%%%%%%%%%%%%%%%%%
%%%%   Table: netD
%%%%%%%%%%%%%%%%%%%%%%%%%%%%%%%%%%%%%%%

\begin{table}[H]
\renewcommand{\arraystretch}{1.3}
\caption{The architecture of the discriminative network. ``IN'' represents InstanceNorm and ``LReLU''
donates leaky ReLU with the slope of 0.2.}
\vspace{-0.5em}
\centering
\begin{tabular}{l}
\hline
\ \ \ \ \  \ \ \  {\bf The architecture of discriminative model} $D$ \ \ \ \ \ \  \ \\
\hline
{\bf Input}: Image ($256 \times 256 \times 3$) \\
\hline
[layer 1]  \ \  \   Conv. (4, 4, \ \ 64), stride=2; \emph{LReLU}; \\
\hline
[layer 2] \   \ \  Conv. (4, 4, 128), stride=2; IN; \emph{LReLU};   \\
\hline
[layer 3] \  \ \    Conv. (4, 4, 256), stride=2; IN; \emph{LReLU};   \\
\hline
[layer 4]  \ \ \  Conv. (4, 4, 512), stride=1; IN; \emph{LReLU};\\
\hline
[layer 5] \  \  \    Conv. (4, 4, 1), stride=1; \emph{Sigmoid};    \\
\hline
{\bf Output}: Real or Fake ($30 \times 30 \times 1$)\\
\hline
\end{tabular}
\label{table:netD}
\end{table}

%%%%%%%%%%%%%%%%%%%%%%%%%%%%%%%%%%%
%% Figure.more_comparison_on_paris_1
%%%%%%%%%%%%%%%%%%%%%%%%%%%%%%%%%%%

%%%%%%%%%%%%%%%%%%%%%%%%%%%%%%
%% Figure.comparison_on_paris_2
%%%%%%%%%%%%%%%%%%%%%%%%%%%%%

%%%%%%%%%%%%%%%%%%%%%%%%%%%%%%%%%%%%%%
%% Figure.more_comparison_on_Places: 1
%%%%%%%%%%%%%%%%%%%%%%%%%%%%%%%%%%%%%%

\clearpage
\subsection{More object removal on real world images by our Shift-Net}
We apply our model trained on Paris StreetView~\cite{doersch2012makes} or Places~\cite{zhou2017places} to process object removal on real world images, as shown in Fig.~\ref{fig:realImgs} for results.
These real world images are complex for large area of distractors and complicated background.
Even so, our model can handle them well, which indicates the effectiveness, applicability and generality of our model.

%%%%%%%%%%%%%%%%%%%%%%%%%%%%%%%%%%%%%%%%%%%%%%
%Figure.Comparisons on real images.(two images)
%%%%%%%%%%%%%%%%%%%%%%%%%%%%%%%%%%%%%%%%%%%%%%
\begin{figure}[!h]
\setlength\tabcolsep{1.5pt}
\centering
\begin{tabular}{cccc}

\includegraphics[width=.24\linewidth]{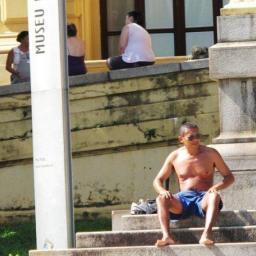}&
\includegraphics[width=.24\linewidth]{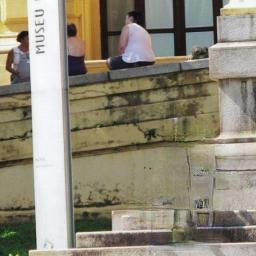}&
\includegraphics[width=.24\linewidth]{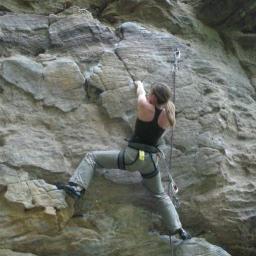}&
\includegraphics[width=.24\linewidth]{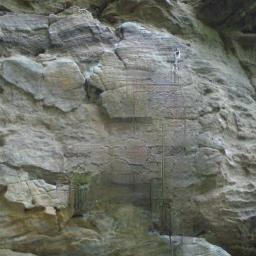}\\
\includegraphics[width=.24\linewidth]{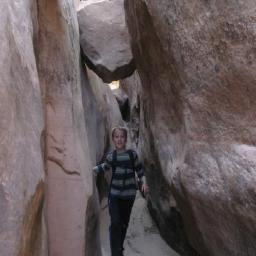}&
\includegraphics[width=.24\linewidth]{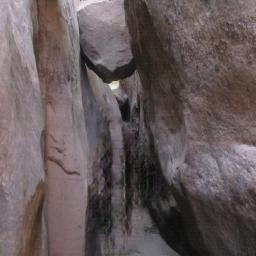}&
\includegraphics[width=.24\linewidth]{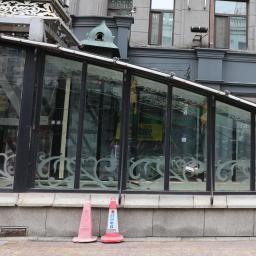}&
\includegraphics[width=.24\linewidth]{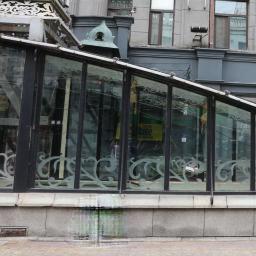}\\

\includegraphics[width=.24\linewidth]{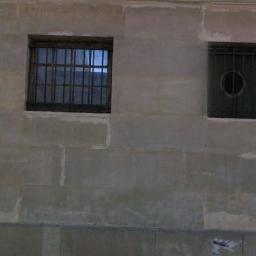}&
\includegraphics[width=.24\linewidth]{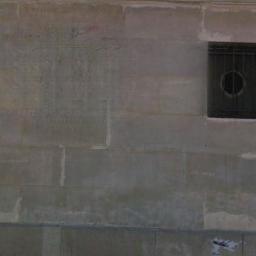}&
\includegraphics[width=.24\linewidth]{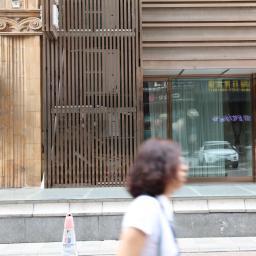}&
\includegraphics[width=.24\linewidth]{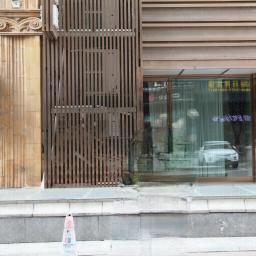}\\

\end{tabular}
\caption{Object removal on real images.}
\label{fig:realImgs}
\end{figure}

\clearpage
\bibliographystyle{splncs}
\bibliography{egbib}

\begin{thebibliography}{10}

\bibitem{Content-Aware-Fill}
Goldman, D., Shechtman, E., Barnes, C., Belaunde, I., Chien, J.:
\newblock Content-aware fill.
\newblock \url{https://research.adobe.com/project/content-aware-fill}

\bibitem{pathak2016context}
Pathak, D., Krahenbuhl, P., Donahue, J., Darrell, T., Efros, A.A.:
\newblock Context encoders: Feature learning by inpainting.
\newblock In: Proceedings of the IEEE Conference on Computer Vision and Pattern
  Recognition. (2016)  2536--2544

\bibitem{barnes2009patchmatch}
Barnes, C., Shechtman, E., Finkelstein, A., Goldman, D.B.:
\newblock Patchmatch: a randomized correspondence algorithm for structural
  image editing.
\newblock In: ACM Transactions on Graphics (TOG). Volume~28., ACM (2009) ~24

\bibitem{yang2017high}
Yang, C., Lu, X., Lin, Z., Shechtman, E., Wang, O., Li, H.:
\newblock High-resolution image inpainting using multi-scale neural patch
  synthesis.
\newblock In: The IEEE Conference on Computer Vision and Pattern Recognition
  (CVPR). (July 2017)

\bibitem{criminisi2003object}
Criminisi, A., Perez, P., Toyama, K.:
\newblock Object removal by exemplar-based inpainting.
\newblock In: Computer Vision and Pattern Recognition, 2003. Proceedings. 2003
  IEEE Computer Society Conference on. Volume~2., IEEE (2003)  II--II

\bibitem{le2011examplar}
Le~Meur, O., Gautier, J., Guillemot, C.:
\newblock Examplar-based inpainting based on local geometry.
\newblock In: Image Processing (ICIP), 2011 18th IEEE International Conference
  on, IEEE (2011)  3401--3404

\bibitem{xu2010image}
Xu, Z., Sun, J.:
\newblock Image inpainting by patch propagation using patch sparsity.
\newblock IEEE transactions on image processing \textbf{19}(5) (2010)
  1153--1165

\bibitem{doersch2012makes}
Doersch, C., Singh, S., Gupta, A., Sivic, J., Efros, A.:
\newblock What makes paris look like paris?
\newblock ACM Transactions on Graphics \textbf{31}(4) (2012)

\bibitem{zhou2017places}
Zhou, B., Lapedriza, A., Khosla, A., Oliva, A., Torralba, A.:
\newblock Places: A 10 million image database for scene recognition.
\newblock IEEE Transactions on Pattern Analysis and Machine Intelligence (2017)

\bibitem{simonyan2014very}
Simonyan, K., Zisserman, A.:
\newblock Very deep convolutional networks for large-scale image recognition.
\newblock arXiv preprint arXiv:1409.1556 (2014)

\bibitem{barnes2010generalized}
Barnes, C., Shechtman, E., Goldman, D.B., Finkelstein, A.:
\newblock The generalized patchmatch correspondence algorithm.
\newblock In: European Conference on Computer Vision, Springer (2010)  29--43

\bibitem{drori2003fragment}
Drori, I., Cohen-Or, D., Yeshurun, H.:
\newblock Fragment-based image completion.
\newblock In: ACM Transactions on graphics (TOG). Volume~22., ACM (2003)
  303--312

\bibitem{efros1999texture}
Efros, A.A., Leung, T.K.:
\newblock Texture synthesis by non-parametric sampling.
\newblock In: Computer Vision, 1999. The Proceedings of the Seventh IEEE
  International Conference on. Volume~2., IEEE (1999)  1033--1038

\bibitem{jia2003image}
Jia, J., Tang, C.K.:
\newblock Image repairing: Robust image synthesis by adaptive nd tensor voting.
\newblock In: Computer Vision and Pattern Recognition, 2003. Proceedings. 2003
  IEEE Computer Society Conference on. Volume~1., IEEE (2003)  I--I

\bibitem{jia2004inference}
Jia, J., Tang, C.K.:
\newblock Inference of segmented color and texture description by tensor
  voting.
\newblock IEEE Transactions on Pattern Analysis and Machine Intelligence
  \textbf{26}(6) (2004)  771--786

\bibitem{komodakis2006image}
Komodakis, N.:
\newblock Image completion using global optimization.
\newblock In: Computer Vision and Pattern Recognition, 2006 IEEE Computer
  Society Conference on. Volume~1., IEEE (2006)  442--452

\bibitem{komodakis2007image}
Komodakis, N., Tziritas, G.:
\newblock Image completion using efficient belief propagation via priority
  scheduling and dynamic pruning.
\newblock IEEE Transactions on Image Processing \textbf{16}(11) (2007)
  2649--2661

\bibitem{pritch2009shift}
Pritch, Y., Kav-Venaki, E., Peleg, S.:
\newblock Shift-map image editing.
\newblock In: Computer Vision, 2009 IEEE 12th International Conference on, IEEE
  (2009)  151--158

\bibitem{simakov2008summarizing}
Simakov, D., Caspi, Y., Shechtman, E., Irani, M.:
\newblock Summarizing visual data using bidirectional similarity.
\newblock In: Computer Vision and Pattern Recognition, 2008. CVPR 2008. IEEE
  Conference on, IEEE (2008)  1--8

\bibitem{sun2005image}
Sun, J., Yuan, L., Jia, J., Shum, H.Y.:
\newblock Image completion with structure propagation.
\newblock ACM Transactions on Graphics (ToG) \textbf{24}(3) (2005)  861--868

\bibitem{wexler2004space}
Wexler, Y., Shechtman, E., Irani, M.:
\newblock Space-time video completion.
\newblock In: Computer Vision and Pattern Recognition, 2004. CVPR 2004.
  Proceedings of the 2004 IEEE Computer Society Conference on. Volume~1., IEEE
  (2004)  I--I

\bibitem{wexler2007space}
Wexler, Y., Shechtman, E., Irani, M.:
\newblock Space-time completion of video.
\newblock IEEE Transactions on pattern analysis and machine intelligence
  \textbf{29}(3) (2007)

\bibitem{kohler2014mask}
K{\"o}hler, R., Schuler, C., Sch{\"o}lkopf, B., Harmeling, S.:
\newblock Mask-specific inpainting with deep neural networks.
\newblock In: German Conference on Pattern Recognition, Springer (2014)
  523--534

\bibitem{ren2015shepard}
Ren, J.S., Xu, L., Yan, Q., Sun, W.:
\newblock Shepard convolutional neural networks.
\newblock In: Advances in Neural Information Processing Systems. (2015)
  901--909

\bibitem{xie2012image}
Xie, J., Xu, L., Chen, E.:
\newblock Image denoising and inpainting with deep neural networks.
\newblock In: Advances in Neural Information Processing Systems. (2012)
  341--349

\bibitem{yeh2017semantic}
Yeh, R.A., Chen, C., Lim, T.Y., Schwing, A.G., Hasegawa-Johnson, M., Do, M.N.:
\newblock Semantic image inpainting with deep generative models.
\newblock In: Proceedings of the IEEE Conference on Computer Vision and Pattern
  Recognition. (2017)  5485--5493

\bibitem{IizukaSIGGRAPH2017}
Iizuka, S., Simo-Serra, E., Ishikawa, H.:
\newblock {Globally and Locally Consistent Image Completion}.
\newblock ACM Transactions on Graphics (Proc. of SIGGRAPH 2017) \textbf{36}(4)
  (2017)  107:1--107:14

\bibitem{li2017generative}
Li, Y., Liu, S., Yang, J., Yang, M.H.:
\newblock Generative face completion.
\newblock arXiv preprint arXiv:1704.05838 (2017)

\bibitem{chen2016fast}
Chen, T.Q., Schmidt, M.:
\newblock Fast patch-based style transfer of arbitrary style.
\newblock arXiv preprint arXiv:1612.04337 (2016)

\bibitem{dumoulin2016learned}
Dumoulin, V., Shlens, J., Kudlur, M.:
\newblock A learned representation for artistic style.
\newblock arXiv preprint arXiv:1610.07629 (2016)

\bibitem{gatys2015neural}
Gatys, L.A., Ecker, A.S., Bethge, M.:
\newblock A neural algorithm of artistic style.
\newblock arXiv preprint arXiv:1508.06576 (2015)

\bibitem{gatys2016controlling}
Gatys, L.A., Ecker, A.S., Bethge, M., Hertzmann, A., Shechtman, E.:
\newblock Controlling perceptual factors in neural style transfer.
\newblock arXiv preprint arXiv:1611.07865 (2016)

\bibitem{huang2017arbitrary}
Huang, X., Belongie, S.:
\newblock Arbitrary style transfer in real-time with adaptive instance
  normalization.
\newblock arXiv preprint arXiv:1703.06868 (2017)

\bibitem{johnson2016perceptual}
Johnson, J., Alahi, A., Fei-Fei, L.:
\newblock Perceptual losses for real-time style transfer and super-resolution.
\newblock In: European Conference on Computer Vision, Springer (2016)  694--711

\bibitem{li2016combining}
Li, C., Wand, M.:
\newblock Combining markov random fields and convolutional neural networks for
  image synthesis.
\newblock In: Proceedings of the IEEE Conference on Computer Vision and Pattern
  Recognition. (2016)  2479--2486

\bibitem{luan2017deep}
Luan, F., Paris, S., Shechtman, E., Bala, K.:
\newblock Deep photo style transfer.
\newblock arXiv preprint arXiv:1703.07511 (2017)

\bibitem{ulyanov2016texture}
Ulyanov, D., Lebedev, V., Vedaldi, A., Lempitsky, V.S.:
\newblock Texture networks: Feed-forward synthesis of textures and stylized
  images.
\newblock In: ICML. (2016)  1349--1357

\bibitem{ronneberger2015u}
Ronneberger, O., Fischer, P., Brox, T.:
\newblock U-net: Convolutional networks for biomedical image segmentation.
\newblock In: Medical Image Computing and Computer-Assisted Intervention
  (MICCAI). (2015)

\bibitem{isola2016image}
Isola, P., Zhu, J.Y., Zhou, T., Efros, A.A.:
\newblock Image-to-image translation with conditional adversarial networks.
\newblock arXiv preprint arXiv:1611.07004 (2016)

\bibitem{zhu2017unpaired}
Zhu, J.Y., Park, T., Isola, P., Efros, A.A.:
\newblock Unpaired image-to-image translation using cycle-consistent
  adversarial networks.
\newblock arXiv preprint arXiv:1703.10593 (2017)

\bibitem{mahendran2015understanding}
Mahendran, A., Vedaldi, A.:
\newblock Understanding deep image representations by inverting them.
\newblock In: Proceedings of the IEEE conference on computer vision and pattern
  recognition. (2015)  5188--5196

\bibitem{ledig2016photo}
Ledig, C., Theis, L., Husz{\'a}r, F., Caballero, J., Cunningham, A., Acosta,
  A., Aitken, A., Tejani, A., Totz, J., Wang, Z.,  et~al.:
\newblock Photo-realistic single image super-resolution using a generative
  adversarial network.
\newblock arXiv preprint arXiv:1609.04802 (2016)

\bibitem{radford2015unsupervised}
Radford, A., Metz, L., Chintala, S.:
\newblock Unsupervised representation learning with deep convolutional
  generative adversarial networks.
\newblock arXiv preprint arXiv:1511.06434 (2015)

\bibitem{kingma2015adam}
Kingma, D.P., Ba, J.L.:
\newblock Adam: A method for stochastic optimization.
\newblock international conference on learning representations (2015)

\bibitem{ulyanov2016instance}
Ulyanov, D., Vedaldi, A., Lempitsky, V.:
\newblock Instance normalization: The missing ingredient for fast stylization.
\newblock arXiv preprint arXiv:1607.08022 (2016)

\end{thebibliography}
\end{document}